\documentclass{article}
\PassOptionsToPackage{numbers, compress}{natbib}
\usepackage[final]{neurips_2025}

\usepackage{microtype}      
\usepackage{xcolor}         

\usepackage[utf8]{inputenc} 
\usepackage[T1]{fontenc}    
\usepackage{hyperref}       
\usepackage{url}            
\usepackage{booktabs}       
\usepackage{amsfonts}       
\usepackage{afterpage}
\usepackage{nicefrac}       
\usepackage{microtype}      
\usepackage{comment}        
\usepackage{amsmath}
\usepackage{multirow}
\usepackage{bm}
\usepackage{amsthm}
\usepackage{hhline}
\usepackage{amssymb}
\usepackage{makecell}
\usepackage{mathtools}
\usepackage{color}
\usepackage{xcolor}
\usepackage{MnSymbol}
\usepackage{graphicx}
\usepackage{arydshln}
\usepackage{booktabs}
\usepackage{framed}
\usepackage[font=small]{caption}
\usepackage[scientific-notation=true]{siunitx}
\usepackage{wrapfig}
\usepackage{lipsum}
\usepackage{sidecap}
\usepackage{pifont}
\usepackage[flushleft]{threeparttable}
\usepackage{diagbox}
\usepackage{enumitem}
\usepackage{tabularx}
\usepackage{nicefrac}       
\usepackage[parfill]{parskip}
\usepackage{bbm}
\usepackage{cases}
\usepackage{subcaption}
\usepackage{algorithm}
\usepackage{longtable}
\usepackage{algorithmic}
\usepackage{hhline}
\usepackage{xspace}
\usepackage{enumitem}
\usepackage{tikz}
\usepackage{todonotes}
\usepackage{graphicx}
\definecolor{backcolour}{rgb}{0.97,0.97,0.97}
\usepackage{listings}
\lstdefinestyle{markdownstyle}{
    basicstyle=\ttfamily\tiny,
    backgroundcolor=\color{backcolour},   
    xleftmargin=0.05\textwidth,
    xrightmargin=0.05\textwidth,
    breakindent=0\dimen0,
    columns=flexible,
    showspaces=false,
    showstringspaces=false,
    breaklines=true,
    breakatwhitespace=true,
    breakautoindent=true,
}
\usepackage[most]{tcolorbox}

\newcommand{\data}{\textsc{MimeQA}}

\newcommand{\qwen}{Qwen2.5-VL}
\newcommand{\llava}{LLaVA-Video}
\newcommand{\gpt}{GPT-4o}
\newcommand{\gemini}{Gemini-1.5-Pro}
\newcommand{\gemininew}{Gemini-2.5-Pro}
\newcommand{\internvl}{InternVL2.5}
\newcommand{\videollama}{VideoLLaMA3}

\NewDocumentCommand{\yi}
{ mO{} }{\textcolor{blue}{\textsuperscript{\textit{May}}\textsf{\textbf{\small[#1]}}}}

\newcommand{\huggingface}{\raisebox{-1.5pt}{\includegraphics[height=1.05em]{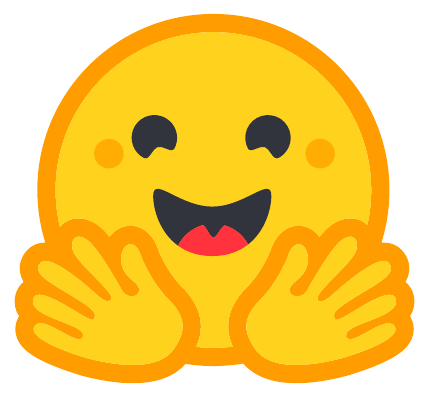}}\xspace}
\newcommand{\github}{\raisebox{-1.5pt}{\includegraphics[height=1.05em]{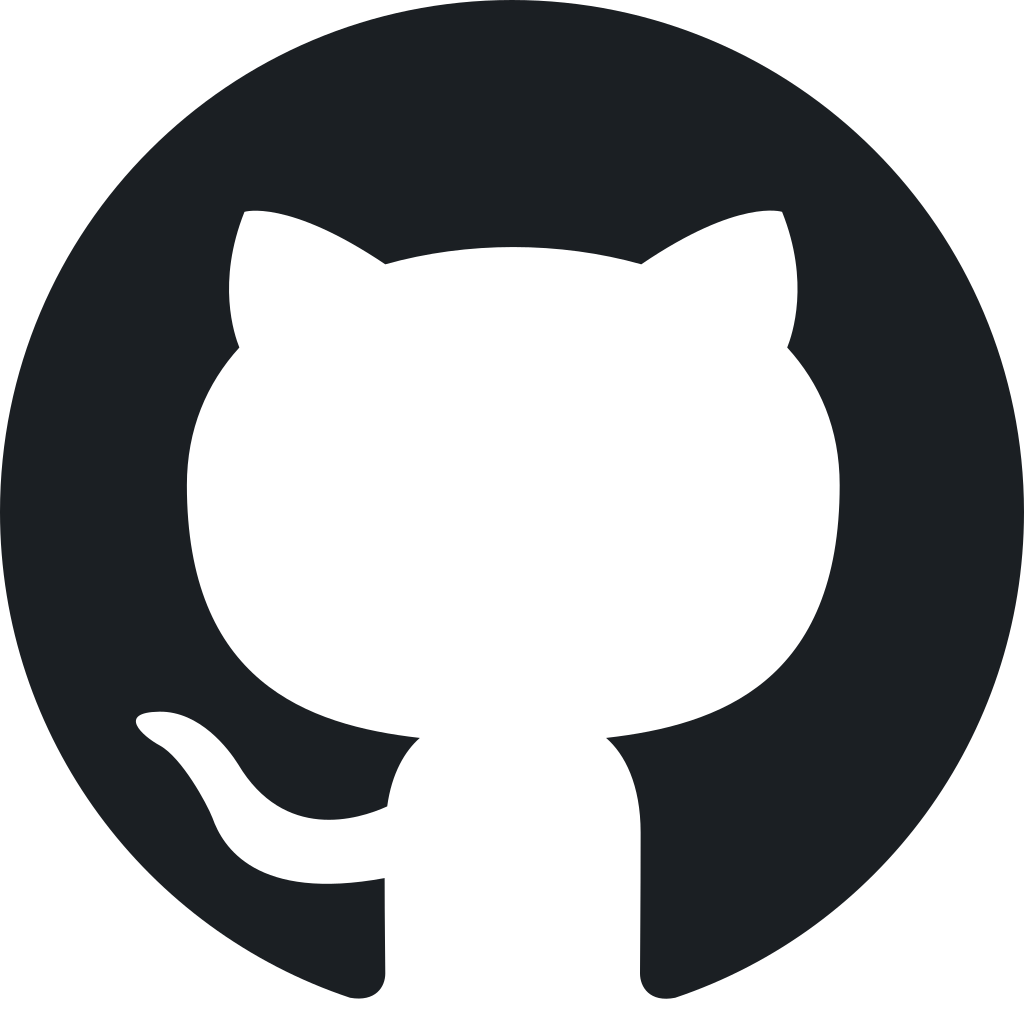}}\xspace}

\hypersetup{
    colorlinks = true,
    linkbordercolor = {white},
    linkcolor = blue!60!black,
    citecolor = blue!60!black,
    urlcolor = blue!60!black
}

\title{\data: Towards Socially-Intelligent Nonverbal Foundation Models}

\author{Hengzhi Li$^{1, 2}$ ~Megan Tjandrasuwita$^{1}$ ~Yi R. Fung$^{1}$\\
\textbf{~Armando Solar-Lezama$^{1}$ ~Paul Pu Liang$^{1}$} \\
$^{1}$Massachusetts Institute of Technology\ ~~~$^{2}$Imperial College London\\
\huggingface \textbf{Data:} \url{https://huggingface.co/datasets/hzli1202/MimeQA} \\
\github \textbf{Code:} \url{https://github.com/MIT-MI/MimeQA}
}

\begin{document}

\maketitle

\vspace{-4mm}

\begin{abstract}

As AI becomes more closely integrated with peoples' daily activities, socially intelligent AI that can understand and interact seamlessly with humans in daily lives is increasingly important. However, current works in AI social reasoning all rely on language-only or language-dominant approaches to benchmark and training models, resulting in systems that are improving in verbal communication but struggle with nonverbal social understanding. To address this limitation, we tap into a novel data source rich in nonverbal social interactions -- mime videos. Mimes refer to the art of expression through gesture and movement without spoken words, which presents unique challenges and opportunities in interpreting nonverbal social communication. We contribute a new dataset called \data, obtained by sourcing \textasciitilde8 hours of videos clips from YouTube and developing a comprehensive video question-answering benchmark comprising 806 carefully annotated and verified question-answer pairs, designed to probe nonverbal social reasoning capabilities. Using \data, we evaluate state-of-the-art video large language models (VideoLLMs) and find that they achieve low accuracy, generally ranging from 20-30\%, while humans score 86\%. Our analysis reveals that VideoLLMs often fail to ground imagined objects and over-rely on the text prompt while ignoring subtle nonverbal interactions. We hope to inspire future work in AI models that embody true social intelligence capable of interpreting non-verbal human interactions. 
\end{abstract}

\section{Introduction}

\begin{figure}[t!]
    \centering
    \vspace{-4mm}
    \includegraphics[width=0.7\linewidth]{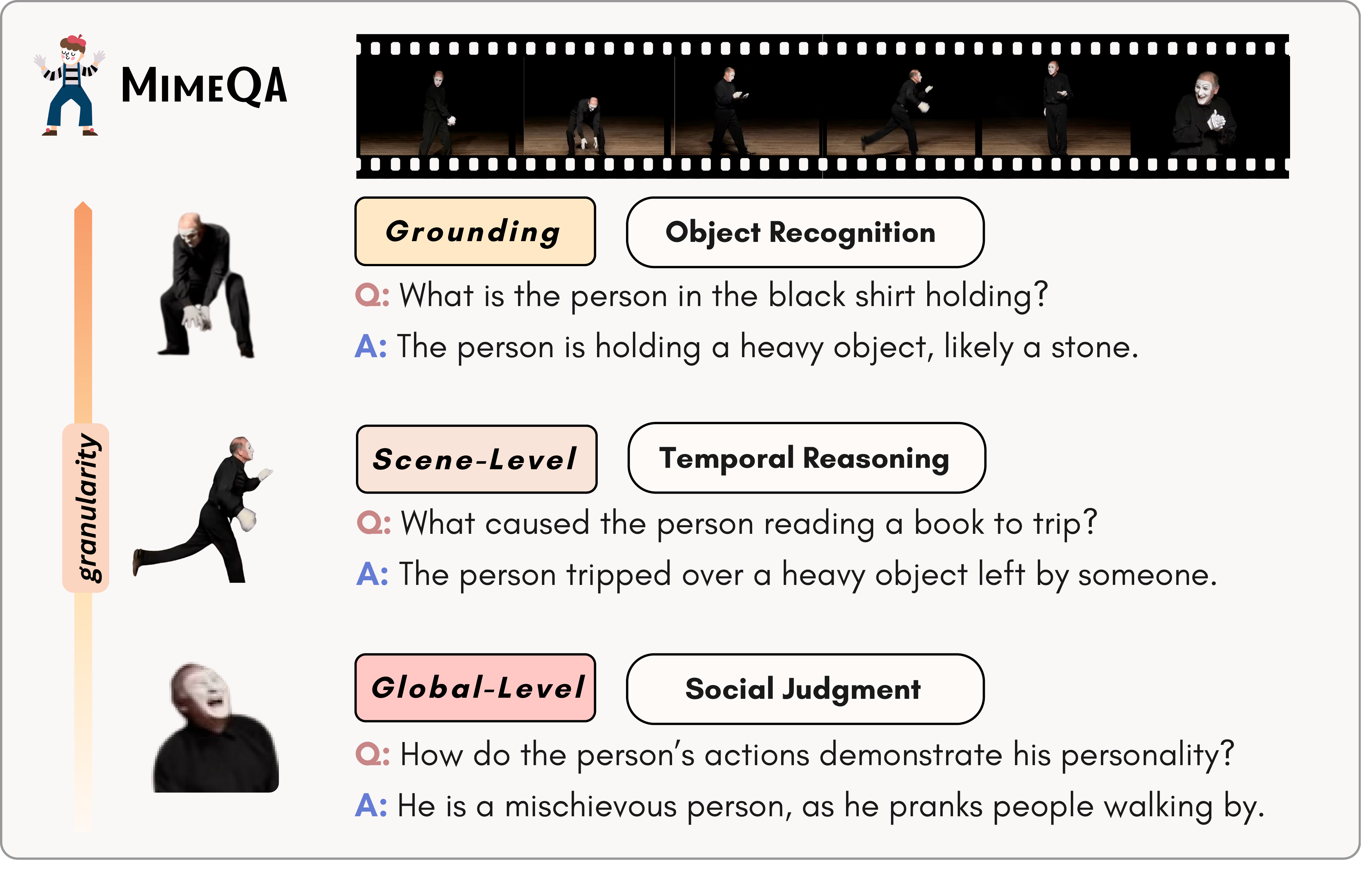}
    \caption{\textbf{\data} is a new benchmark testing nonverbal social reasoning in multimodal large language models, with 101 videos of mimes (the art of expression through gesture without spoken words), and 806 question-answer pairs at three levels: 1) grounding the imagined object or activity, 2) scene-level understanding, and 3) global-level questions on holistic social comprehension. Most models achieve only 20-30\% accuracy.}
    \vspace{-4mm}
    \label{fig:overview}
\end{figure}

Social intelligence is integral to human interactions and enables nuanced understanding and communication with others \cite{mathur-etal-2024-advancing,gweon2023socially,breazeal2003toward}. There is increasing interest in developing socially intelligent AI systems that can understand and interact seamlessly with humans to help them in daily lives, such as stimulating empathic conversations in online mental health forums \cite{sharma2023human}, assisting patients in geriatric care \cite{gonzalez2021social, fleming2003caregiver}, supporting children with autism spectrum conditions \citep{hurst2020social, scassellati2012robots}, and helping educators in classroom teaching \cite{woo2021use}. However, the majority of research towards socially intelligent AI focuses on language-only data and tasks (e.g., question-answering and dialogue)~\citep{kim2023soda,sap2019socialiqa}, or multimodal data where language is often primary and nonverbal modalities (e.g., vocal and visual expression) are treated as second-class citizens~\citep{liang2024foundations,zadeh2018multimodal}. This results in a fundamental mismatch where today's foundation models are strong at language understanding but have a generally poor command of nonverbal social interactions; for example, nonverbal theory-of-mind~\cite{kampisNonverbalComponentsTheory2017}, facial expression~\cite{huang2023language,liang2024hemm}, group social dynamics~\cite{shum2019theory}, and egocentric goal-oriented reasoning~\cite{jiaEgoTaskQAUnderstandingHuman2022} are all challenges for today's language and multimodal foundation models.

To address these limitations, we tap into a novel data source rich in nonverbal social interactions -- \textbf{mime performances}. Mimes refer to the art of expression through gesture and movement without spoken word \cite{zywiczynski2018defining}, which presents unique challenges and opportunities for AI \cite{phutela2015importance}. Since mime performances are devoid of speech, props, and actual objects, instead relying solely on the mime's ability to convey messages, emotions, and narratives through nonverbal communication, AI models must have an acute understanding of human behavior, theory of mind, and the `imagined' objects and actions they convey. 
Furthermore, mimes often depict complex interpersonal relationships and affective states from nonverbal interactions alone, without explicit narration and dialogue.

To systematically assess proficiency on these tasks, we create a benchmark called \data, obtained by sourcing 221 mime videos spanning 8 hours of content from Youtube, annotating each video with questions ranging from local grounding tasks to broader theory of mind and social norm understanding, and meticulous verification of the annotations, resulting in 101 videos and 806 QA pairs. We benchmark state-of-the-art open-source and closed-source VideoLLMs and find that the overall accuracy ranges mostly from 20\% to 30\%, while humans perform 86\%. Our extensive error analysis and ablations point to fundamental shortcomings of VideoLLMs' social and visual understanding capabilities, as common failure modes include failing to recognize imagined objects, misinterpreting nuanced social cues, and hallucinating responses based on the text input. We release our benchmark and evaluation framework at \url{https://github.com/MIT-MI/MimeQA} to drive future research toward verbal and nonverbal social intelligence in AI systems.

\section{\mbox{Theoretical Grounding \& Related Work}}
\begin{figure*}
    \centering
    \vspace{-4mm}
    \includegraphics[width=\linewidth]{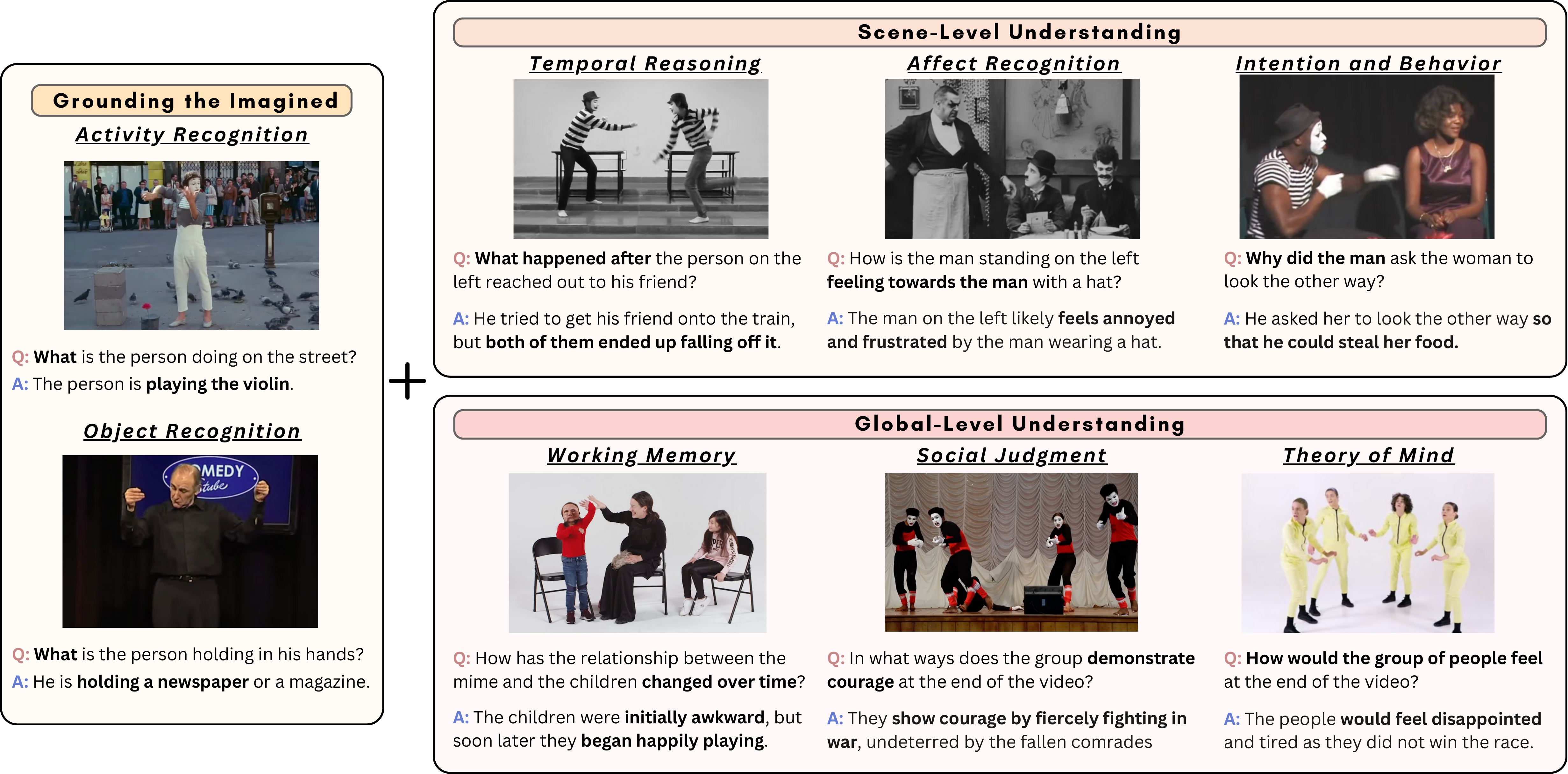}
    \vspace{-1mm}
    \caption{\textbf{Examples of {\data} question types.} \textbf{Left:} Grounding the imagined questions includes recognizing the activity or pretend object that the mime is acting out. \textbf{Top right:} Scene-level questions include temporal reasoning about a localized sequence of events, affect recognition questions about the emotional state of the characters, and intention and behavior questions that require interpreting the goals and motivations within a scene. \textbf{Bottom right:} Global-level questions involve working memory questions that probe understanding of the plot beyond localized sequences, social judgment questions about how the characters' actions adhere to cultural and social norms, and theory of mind questions about the characters' beliefs, desires, and motivation.}
    \vspace{-4mm}
    \label{fig:question_examples}
\end{figure*}

Building \textbf{socially intelligent AI} involves creating agents that can sense, perceive, reason about, learn from, and respond to the affect, behavior, and cognition of other agents (human or artificial), and is a key part of AI as it becomes increasingly involved in our everyday lives~\citep{mathur-etal-2024-advancing}. To push the frontiers of socially intelligent AI, a rich body of work has examined various modalities, including language, video, audio, and more. For example,~\citet{gandhiUnderstandingSocialReasoning2023} evaluates the capabilities of AI to model human mental states from language to predict human goals and future behavior. Related work has also focused on extracting fine-grained visual features from gaze~\citep{singhCombiningGazeAI2020, zhangHumanGazeAssisted2020}, expressions~\citep{zheng-etal-2023-facial, zheng2024unimodal}, and body language~\citep{xu-etal-2024-llm,ozaki2024bqabodylanguagequestion,yoon2019robots, liu2022learning}. Multimodal approaches have also been proposed to provide more holistic human intent understanding~\citep{liang2024foundations,liang2024hemm,mathur2025social}.~\citet{siq2} evaluate video understanding of social situation via question-answering,~\citet{jin-etal-2024-mmtom} evaluate theory of mind question answering on human household activities, and~\citet{li2023intentqa} evaluate human intent understanding in videos. 

Recent advances in \textbf{large multimodal models} have shown impressive video understanding capabilities in various domains, such as egocentric understanding and navigation \citep{mangalam2023egoschema}, multimedia content analysis \citep{liVideoVistaVersatileBenchmark2024a}, and human language understanding~\citep{liang2024hemm,tsai2019multimodal}. State-of-the-art enterprise models, such as Google Gemini \citep{team2024gemini} and GPT-4 \citep{achiam2023gpt}, and open-source models such as Qwen-VL 
\citep{qwen2.5-VL} and LLaVA-Video \citep{zhang2024video} have long context windows capable of handling video and audio inputs. These multimodal models have significantly improved performance on recent challenging video question-answering benchmarks \citep{nagrani2024neptune, mangalam2023egoschema, rawal2024cinepile, fu2024video}. Despite significant progress, most existing models rely primarily on the language modality~\citep{liang2024hemm}, resulting in commonsense biases in question prompts and, in extreme cases, good performance even without access to video at all \citep{min2024morevqa}. Consequently, there is a lack of benchmarks that effectively evaluate the social intelligence of AI beyond language.

\textbf{Mime performances} serve as a good case for measuring nonverbal social intelligence. Mimes, or pantomimes when the performance has a coherent narrative, are often considered a peripheral form of communication due to their independence of speech and lack of structured conventions \citep{mcneill2008gesture, mcneill2012language}. Nevertheless, pantomimes have a crucial place in developing the human's natural language system; they are often seen as the fundamental building block to human language evolution, where systematic grammatical systems arise from increasingly complex gestural interaction over time \citep{kendon2017reflections, mineiro2017emerging, zlatev2020pantomime, ferretti2023influence}. From a cognitive development perspective, \citet{arbib2017toward, arbib2024pantomime} posits that pantomimic gestures are crucial in the development from ``language-ready'' to ``language use'' brains, and studies have found that pantomime understanding is related to causal reasoning, working memory, and theory of mind capabilities \citep{adornetti2023comprehending, gardenfors2024relations}. In human everyday communication, the highly iconic and transparent nature of pantomimic gestures leads to their frequent use in language-restrained settings, such as language impairment \citep{fex1998use, goldin2005resilience}, cross-cultural communication \citep{ortega2020types, zywiczynski2021pantomimic}, and neurodivergent communication \citep{yavuz2019social}. Thus, mime performance presents a rich and untapped source for benchmark nonverbal social understanding in modern AI systems. While some prior works used mimes for evaluating action recognition \citep{weinzaepfel2021mimetics, cho-etal-2025-vision}, to our best knowledge, \data\ is the first dataset to use mime performances to holistically evaluate multimodal foundation models’ nonverbal social intelligence.

\section{{\data} Dataset}\label{sec:data}

We operationalize the opportunities and challenges of building nonverbal social intelligence through mime videos in a new open-ended video question-answering benchmark called {\data}. This benchmark consists of questions that evaluate social understanding at varying levels, from basic perception to complex reasoning about social dynamics across the full video.

\subsection{Question Hierarchy}
The {\data} questions are structured into three levels across the temporal scale, progressing from low-level visual recognition to scene-level interpretation and global-level cognitive reasoning. See Figure \ref{fig:question_examples} for example questions for each category.

\paragraph{Grounding the Imagined.} An important element of mime performances is its use of abstract iconic gestures or body movements to convey an imagined object or activity \cite{zywiczynski2018defining}. For example, a movement of flapping one's wings may represent a flying bird. These gestures are grounded in humans' embodied experience, and understanding their meaning is crucial for mimic communication \cite{gardenfors2017demonstration, zlatev2020pantomime}. To measure the VideoLLMs' capabilities to ground these imagined objects and actions, our first level of questions involves recognizing basic visual elements in the mime performance, such as objects and activities. This foundational perceptual information is a precursor for higher-level reasoning about interactions and intentions, as shown by \citet{sibierska2022s}.

\paragraph{Scene-Level.} This level moves beyond perception to examine social interactions within a short video segment. Inspired by previous benchmarks \cite{xiao2021next, siq2} and cognitive development research \cite{burris2014all}, we define three categories to assess fine-grained social understanding at the scene level.
\begin{itemize}[noitemsep,topsep=0pt,nosep,leftmargin=*,parsep=0pt,partopsep=0pt]
    \item \textbf{Temporal reasoning} \cite{trabasso1989logical} requires structuring events into a causal chain linked by logical necessity and transitivity. This category involves identifying sequences of events in a scene and their temporal-causal relationships, beyond mere event ordering.
    \item \textbf{Affect recognition} \cite{pantic2003toward} involves identifying and analyzing emotional states through nonverbal cues. Other than static emotion classification, this category also requires detecting subtle emotional shifts, group sentiment, and changes in expression.
    \item \textbf{Intention and behavior understanding} \cite{blakemore2001perception} involves inferring the motivations behind actions and interpreting how observed behavior reflects unobserved internal goals and mental states.
\end{itemize}

\paragraph{Global-Level.} This level assesses the ability to synthesize and reason social information across multiple scenes. Unlike scene-level understanding, it prioritizes organizing and weighing social cues to form higher-order interpretations rather than isolated moments. Drawing from research on non-linguistic narrative comprehension \cite{baron1986mechanical, kuijper2017narrative, adornetti2023comprehending}, we define three categories to evaluate global social intelligence.
\begin{itemize}[noitemsep,topsep=0pt,nosep,leftmargin=*,parsep=0pt,partopsep=0pt]
    \item \textbf{Working memory} \cite{daneman1996working} involves retrieving, integrating, and reasoning information across the entire video. Beyond single events, these questions require the ability to determine the relevance of past information, recall key events, and synthesize a coherent narrative. 
    \item \textbf{Social judgment} \cite{kahneman1986norm} involves evaluating behaviors, assessing personality traits, and identifying social constructs like rapport, trust, and cooperation. This category requires comparing observations to social norms and counterfactual alternatives, highlighting unexpected or abnormal behavior.
    \item \textbf{Theory of mind} \cite{astington1995theory} measures the ability to infer beliefs, goals, and perspectives. This ability enables perspective-taking, reasoning about unseen motives, and anticipating how different individuals understand the same situation.
\end{itemize}

\subsection{Dataset Construction}
\begin{figure}[t]
    \centering
    \vspace{-4mm}
    \includegraphics[width=\linewidth]{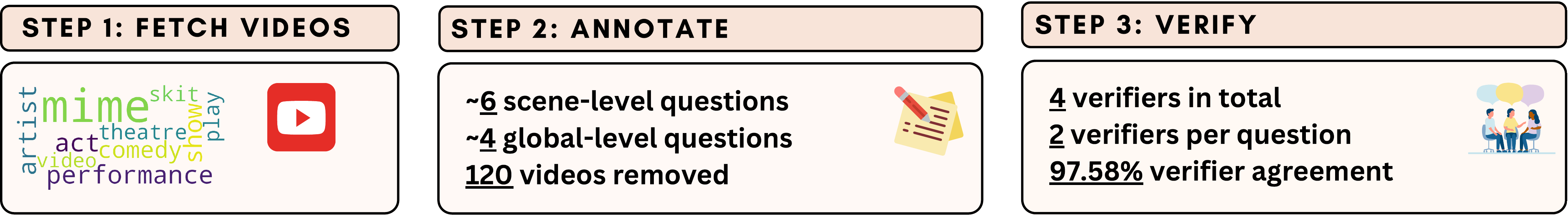}
    \vspace{-2mm}
    \caption{\textbf{Dataset construction pipeline:} 1) Collecting videos from YouTube with various search terms that are summarized by the word cloud. 2) Annotating approximately 6 grounding and scene-level questions and 4 global-level questions per video, removing 120 videos in the process. 3) Verifying the annotated questions and answers, with 97.58\% verifier agreement.}
    \vspace{-2mm}
    \label{fig:dataset_pipeline}
\end{figure}

We summarize our dataset construction pipeline in Figure \ref{fig:dataset_pipeline} and detail individual steps below.

\paragraph{Video collection.} We collect videos from YouTube using various search terms that include the keyword ``mime'', downloading up to 50 videos per keyword. See Figure \ref{fig:dataset_pipeline} for a word cloud of the search terms. We restrict video durations to between one and ten minutes. Additionally, we only select videos licensed under Creative Commons. This process yields a dataset of 221 videos.

\paragraph{Video validation and annotation.} We asked two human annotators familiar with the question hierarchy to generate questions for each video, along with one-sentence answers to the question. The annotators are provided with a comprehensive description of the question hierarchy alongside a few examples per category. To ensure a diversity of categories, for each video, the annotators are asked to annotate approximately six scene-level questions, four global questions, and as many grounding questions as relevant, although the actual number of questions may vary based on the video. For grounding and scene-level questions, we asked them to provide start and end timestamps denoting the segment that the question is referring to. During the annotation process, annotators eliminated videos that lack a plot, are too difficult to understand, or explicitly involve language such as song lyrics or verbal explanations. We use the VGG Image Annotator \cite{dutta2019vgg} for all annotations.

\paragraph{Annotation verification.} After an annotator has created a set of questions and answers for a video, a second person who has not seen the video verifies the quality of the annotation. The verifier is asked to watch the videos, answer the set of questions, and compare their answer with the originally annotated ground truth. The verifier then marks whether the two answers are consistent or otherwise provides suggestions to refine the questions. Finally, we manually review the verification results, remove any questions with inconsistent answers to avoid ambiguity, and refine the questions based on suggestions. By the end of this process, we reduced the original set of questions to 806 questions. See Figure \ref{fig:dataset_pipeline} for an illustration of the dataset construction pipeline.

\subsection{Dataset Statistics}

\begin{figure}[t]
  \vspace{-2mm}
  \centering
  \begin{subfigure}[t!]{0.68\linewidth}
      \includegraphics[width=\linewidth]{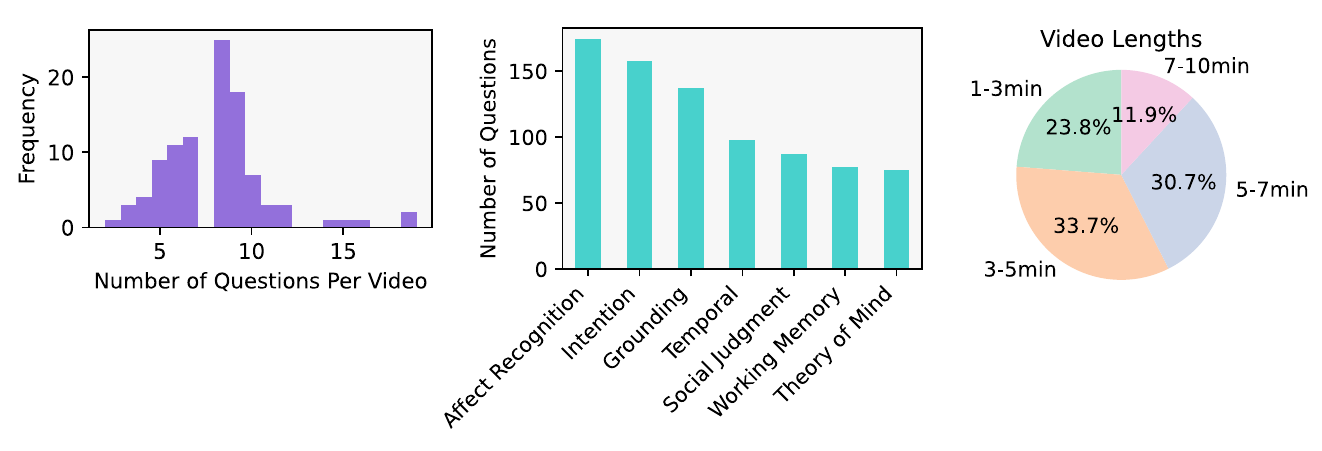}
  \end{subfigure}
  \begin{subfigure}[t!]{0.3\linewidth}
  \centering
  \vspace{-10mm}
  \resizebox{\linewidth}{!}{
        \begin{tabular}{cc}
        \toprule
        \textbf{Statistic} & \textbf{Value}   \\
        \hline
        Avg. \# of Questions per Video &  7.98  \\
        Avg. Video Length & 4.57 min \\
        Total \# of Annotated Questions & 826 \\
        \% of Discarded Questions  & 2.42\%\\
        \% of Modified Questions& 2.49\% \\
        \bottomrule
        \end{tabular}
        \label{tab:statistics}
    }
  \end{subfigure}
  \vspace{-2mm}
  \caption{\textbf{{\data} dataset statistics.} Distribution of video lengths shows the range of short to long timescales. The distribution of the number of questions per video shows that each video is densely annotated, and the distribution of the number of questions per category is balanced.}
  \vspace{-6mm}
  \label{fig:statistics}
\end{figure}

Figure \ref{fig:statistics} contains \data's dataset statistics. The videos are densely annotated, with 806 total questions and most videos having more than five questions. \data\ has balanced questions across categories, with over 70 questions for each global category and over 100 questions for each local category. We provide additional analysis on \data's gender, cultural, and human behavior diversity in Appendix \ref{app:diversity}.
\section{Experiments}
\label{sec:exps}
In this section, we evaluate closed and open-source VideoLLMs on the {\data} dataset. We detail the evaluation setup, present quantitative results, and conduct error analysis to understand model behavior in non-verbal social reasoning.

\subsection{Experimental setup}

We evaluate state-of-the-art closed- and open-source VideoLLMs on {\data} based on performance on video understanding benchmarks \citep{fu2024video, wu2025longvideobench}. For closed-source, we selected \gemini~\citep{team2024gemini}, \gemininew~\citep{comanici2025gemini}, and \gpt~\citep{achiam2023gpt}, and we selected open-source models \qwen~\citep{qwen2.5-VL}, \llava~\citep{zhangVideoInstructionTuning2024}, \internvl~\citep{chenExpandingPerformanceBoundaries2025a}, and \videollama~\citep{zhangVideoLLaMA3Frontier2025}. We use a standardized prompt, where we introduce the task of understanding mimes and subsequently ask a question, potentially including timestamps if it is a grounding or scene-level question. For models that do not natively support video input, we uniformly sample a number of frames with timestamps attached to the frames. See Appendix \ref{app:prompt} for the evaluation prompt template and Appendix \ref{app:model_settings} for model settings.

To evaluate the model accuracy on our open-ended QA task, we use \gpt~for LLM-as-a-judge \citep{zheng2023judging} to automatically verify the model response against ground truth answers. A response is considered correct if it is semantically equivalent to the annotated ground truth. We evaluate the LLM grader quality on a sample of 352 questions and find that the automated grader aligns with a human grader 92.0\% percent of the time. 
See Appendix \ref{app:prompt} for LLM grader prompt.

\subsection{Results}

\begin{table*}[t]
\centering
\caption{\textbf{Model accuracies on vision-language and language-only inputs across different questions.} VL=Video and text, L=Text only. \textbf{Avg}=Average overall performance. \textbf{GI}=Grounding the Imagined, \textbf{I}=Intention, \textbf{AR}=Affect Recognition, \textbf{T}=Temporal, \textbf{ToM}=Theory of Mind, \textbf{SJ}=Social Judgment, \textbf{WM}=Working Memory.}
\scalebox{0.69}{
\begin{tabular}{ccc |cc |cc|cc|cc| cc|cc|cc}
\toprule
\multirow{3}{*}{\textbf{Model}}&   & &  \multicolumn{2}{c}{\textbf{Grounding}} & \multicolumn{6}{|c|}{\textbf{Scene-Level}} & \multicolumn{6}{c}{\textbf{Global-Level}}  \\ \cline{4-17}

& \multicolumn{2}{c}{\textbf{Avg}}  & \multicolumn{2}{|c|}{\textbf{GI}} & \multicolumn{2}{c}{\textbf{I}} & \multicolumn{2}{|c|}{\textbf{AR}} & \multicolumn{2}{c}{\textbf{T}} & \multicolumn{2}{|c}{\textbf{ToM}} & \multicolumn{2}{|c|}{\textbf{SJ}} & \multicolumn{2}{c}{\textbf{WM}} \\  
\cline{4-17}

& VL & L & VL & L  & VL & L  & VL & L & VL  & L & VL  & L & VL  & L & VL  & L \\
\hline
\qwen~\cite{qwen2.5-VL}&  20.1 & 13.2 & 6.6 & \textbf{9.5} & 15.8 & 8.9& 23.6 & 14.4 & 14.3 & 6.1& 38.7 & 29.3 & 33.3 & 18.4 & 19.4 & 13.0  \\
\llava~\cite{zhangVideoInstructionTuning2024} & 19.4 & 17.2 & 9.5 & 7.3 & 13.3 & 13.3& 25.9 & 23.6& 8.2 & 11.2 & 26.7 & 26.7 & 39.1 & 25.3& 19.5 &  18.2 \\
InternVL2.5~\cite{chenExpandingPerformanceBoundaries2025a} & 21.6 & \textbf{20.6}  & 7.3 & 5.8 & 22.2 & \textbf{17.1} & 28.2 & \textbf{25.9} & 15.3 & \textbf{12.2} &32.0 & 32.0 & 33.3 & \textbf{43.7} & 15.6 & 15.6 \\
VideoLLaMA3~\cite{zhangVideoLLaMA3Frontier2025} & 22.2 & 19.5 & 7.3 & \textbf{9.5} & 13.3 & 12.0 & 34.5 & 24.7 & 13.3 & \textbf{12.2} & 41.3 & \textbf{36.0} & 31.0 & 31.0 & 22.1 & \textbf{20.8} \\
\hline
\gpt~\cite{achiam2023gpt} & 31.3 & 18.1 & 19.0 & 6.5 & 28.5  & 14.6 & 29.9 & 22.4 & \textbf{30.6} & 10.2 & 45.3 & 32.0 & 43.7 & 33.3 & 35.1 & 15.6 \\
\gemini~\cite{team2024gemini} & 30.6 & 13.3 & 20.4 & 5.8 & 22.8 & 10.8 & 34.5 & 17.2 & \textbf{30.6} & 8.2 & 42.7 & 20.0 & 40.2 & 23.0 & 33.7 & 11.7  \\
\gemininew~\cite{comanici2025gemini} & \textbf{38.3} & 15.5 & \textbf{28.4} & 4.4 & 
\textbf{31.6} & 10.8 & \textbf{43.7} & 19.5 & 28.6 & \textbf{12.2} & \textbf{54.7} & 
32.0 & \textbf{51.7} & 25.3 & \textbf{39.0} & 13.0  \\
\hline
Human & 86.0 & - & 89.8 & -& 87.3 & -& 83.9 & -& 88.8 & -& 93.3 &-& 80.5 & -& 76.6 & -\\
\bottomrule
\end{tabular}
}
\vspace{-2mm}
\label{tab:models}
\end{table*}

We report the performance of open-source and closed-source models in Table \ref{tab:models}. All models achieved low performance on the dataset: the open-source models achieve approximately 20\% average accuracy, whereas {\gpt} achieves 31.3\% and {\gemininew} obtains 38.3\%, significantly below the human baseline of 86.0\%. This highlights the continued challenge for current models in visual abstraction and recognizing subtle social cues. In general, models perform better on global-level questions than on scene-level and grounding questions, suggesting that models struggle more with fine-grained video understanding compared to grasping the overall context of a video. Notably, models perform especially poorly on the grounding category, indicating a significant limitation in models' abstract visual cognition on imagined objects. {\gpt}, {\gemini}, and {\gemininew} outperform open-source models by a factor of 2–3$\times$ across most categories.

To assess language bias in our dataset, we ablate the effect of video information by evaluating all models on text-only input, excluding video. We observe that models achieve higher accuracy on global-level questions than on scene-level ones without access to video. For example, without video, \internvl~achieved 43.7\% accuracy on the Social Judgment category, but only 5.8\% on Grounding. This bias in global-level questions likely arises because some questions often include additional context to avoid referring to specific video segments, making it easier for models to infer information from annotations alone. Among open-source models, we observe an accuracy drop of 1-7\% when transitioning from video to text-only evaluation. Interestingly, open-source models do not always benefit from video input. For example, both \qwen~and \videollama~observe a drop in accuracy in the Grounding category when provided with video, suggesting they struggle to integrate visual information effectively in question answering. In contrast, {\gpt}, {\gemini}, and {\gemininew} demonstrate significantly better video comprehension, showing substantial accuracy improvements across all categories when provided with video input.

\subsubsection{Improving Model Performance on \data}
In this section, we attempt two distinct approaches to improve model performance on \data: supervised finetuning and the integration of explicit pose-based information. We also experimented with chain-of-thought prompting \citep{wei2022chain} in Appendix \ref{app:cot}, which yielded no gains in accuracy.

First, we evaluated whether supervised finetuning could enhance nonverbal social reasoning. From Table \ref{tab:qwen-ft-compact}, we observe that finetuning the 72B Qwen2.5-VL model \citep{qwen2.5-VL} on 80\% of the \data\ benchmark improved overall accuracy from 22.5\% to 26.6\% on the held-out test set, reaching a level comparable to proprietary models like \gemini. The most significant gains occurred in categories requiring higher-level social reasoning, such as Intention (18.8\% to 28.1\%), Theory of Mind (44.4\% to 55.6\%), and Working Memory (23.5\% to 47.1\%). This suggests finetuning effectively improves the model's comprehension of nonverbal human behavior. However, Grounding performance remained low at 7.1\%, indicating persistent challenges in inferring mimed objects and motivating our next experiment.

\begin{table}[h]
\centering
\vspace{-4mm}
\caption{\textbf{Fine-tuned vs. base {\qwen} on per-category accuracy.} FT = fine-tuned. GI = Grounding the Imagined, I = Intention, AR = Affect, T = Temporal, ToM = Theory of Mind, SJ = Social Judgment, WM = Working Memory. Finetuning {\qwen} on {~\data} improves overall performance on test set.}
\vspace{2mm}
\scalebox{0.85}{
\begin{tabular}{l c | c | c c c | c c c}
\toprule
\textbf{Model} & \textbf{Avg} 
& \multicolumn{1}{c|}{\textbf{Grounding}} 
& \multicolumn{3}{c|}{\textbf{Scene-Level}} 
& \multicolumn{3}{c}{\textbf{Global-Level}} \\
\cmidrule(lr){3-3} \cmidrule(lr){4-6} \cmidrule(lr){7-9}
& & \textbf{GI} & \textbf{I} 
& \textbf{AR} & \textbf{T} & 
\textbf{ToM} & \textbf{SJ} & \textbf{WM} \\
\midrule
\qwen&  22.5  & 7.1  & 18.8 & 16.7  & 17.4& 44.4 & \textbf{47.4}  & 23.5 \\
Qwen2.5-FT & \textbf{26.6} & \textbf{7.1} & \textbf{28.1} & \textbf{19.4} & \textbf{17.4} & \textbf{55.6} & 31.6 & \textbf{47.1}\\
\bottomrule
\end{tabular}
}
\vspace{-2mm}
\label{tab:qwen-ft-compact}
\end{table}

To address the models' persistent low performance in grounding imagined objects and actions, we explored whether incorporating explicit human pose information as input could enhance performance on \data. We used PoseC3D \citep{duan2022revisiting}, a skeleton-based action recognition model trained on the NTU RGB+D \citep{shahroudy2016ntu} dataset, to generate timestamped action labels for a 30\% subset of \data\ at 10-second intervals. We retained only action predictions with confidence scores above 60\%, which is chosen arbitrarily after manually checking a sample of the recognition results. These action labels were then included as additional input to the evaluation prompt of a Qwen2.5-VL-72B model. The prompt explicitly noted that the action recognition results are potentially unreliable.

This approach yielded a discernible trade-off. We saw a marginal decline in overall accuracy from 16.84\% to 16.16\% on this subset of \data. Notably, performance on fine-grained perceptual tasks improved, with accuracy on Grounding the Imagined increased from 2.33\% to 6.98\%, and Working Memory rose from 14.29\% to 23.81\%. Meanwhile, performance on higher-level social inference tasks dropped, with Theory of Mind decreasing from 52.17\% to 34.78\% and Social Judgment from 34.48\% to 24.14\%. We posit that the performance declines can be attributed to model hallucinations induced by inaccurate or overly literal action labels from the perception module, especially since PoseC3D has a limited number of 60 action labels. For example, a quarrel between actors is sometimes misclassified as “nausea or vomiting condition”, which misled the model during reasoning. 

Nevertheless, the improvement in the Grounding category represents a critical gain not achievable through supervised fine-tuning alone. This finding suggests that integrating explicit perception modules is a promising avenue for resolving fundamental grounding challenges in \data. However, future work must develop more sophisticated integration strategies that preserve low-level grounding benefits without compromising the model's capacity for high-level social reasoning.

\subsubsection{Studying Transfer Between \data\ and Other Social Intelligence Tasks}

To assess how well the nonverbal social reasoning skills learned from \data transfer to other tasks, we performed a cross-dataset generalization experiment with two other social intelligence benchmarks: Social-IQ 2.0~\cite{siq2} and IntentQA~\cite{li2023intentqa}. For each source dataset, we conducted five-fold cross-validation, finetuning a 7B \qwen\ model on four folds (80\%) and validating on the remaining fold (20\%) in each split. This allows us to measure both the finetuned model's in-domain performance (e.g., trained on \data, tested on \data) and its ability to generalize to the other two tasks. Since Social-IQ 2.0 and IntentQA are multiple-choice question-answering datasets, we adapt them to open-ended QA by using the correct choice texts as the target answers.

\begin{table}[t]
    \centering
    \caption{\textbf{Cross-dataset generalization results.} We report the average accuracy improvement from a five-fold cross-validation between \data, Social-IQ 2.0, and IntentQA. Models finetuned on MimeQA show good generalization to Social-IQ 2.0 and IntentQA, while models trained on other datasets struggle on \data.}
    \vspace{1mm}
    \scalebox{0.9}{
        \begin{tabular}{lccc}
        \toprule
         & \textbf{MimeQA Test} & \textbf{Social-IQ Test} & \textbf{IntentQA Test} \\
        \midrule
        Finetuned on MimeQA & \textbf{3.5\% $\pm$ 3.1\%} & \textbf{1.2\% $\pm$ 3.0\%} & 2.6\% $\pm$ 1.6\% \\
        Finetuned on Social-IQ 2.0 \citep{siq2} & 0.4\% $\pm$ 2.3\% & 1.0\% $\pm$ 2.3\% & N/A \\
        Finetuned on IntentQA \citep{li2023intentqa} & 1.1\% $\pm$ 3.4\% & N/A & \textbf{3.7\% $\pm$ 1.2\%} \\
        \bottomrule
        \end{tabular}
    }
    \vspace{-2mm}
    \label{tab:transfer_cv}
\end{table}

\begin{figure}[t]
    \centering
    \includegraphics[width=0.95\linewidth]{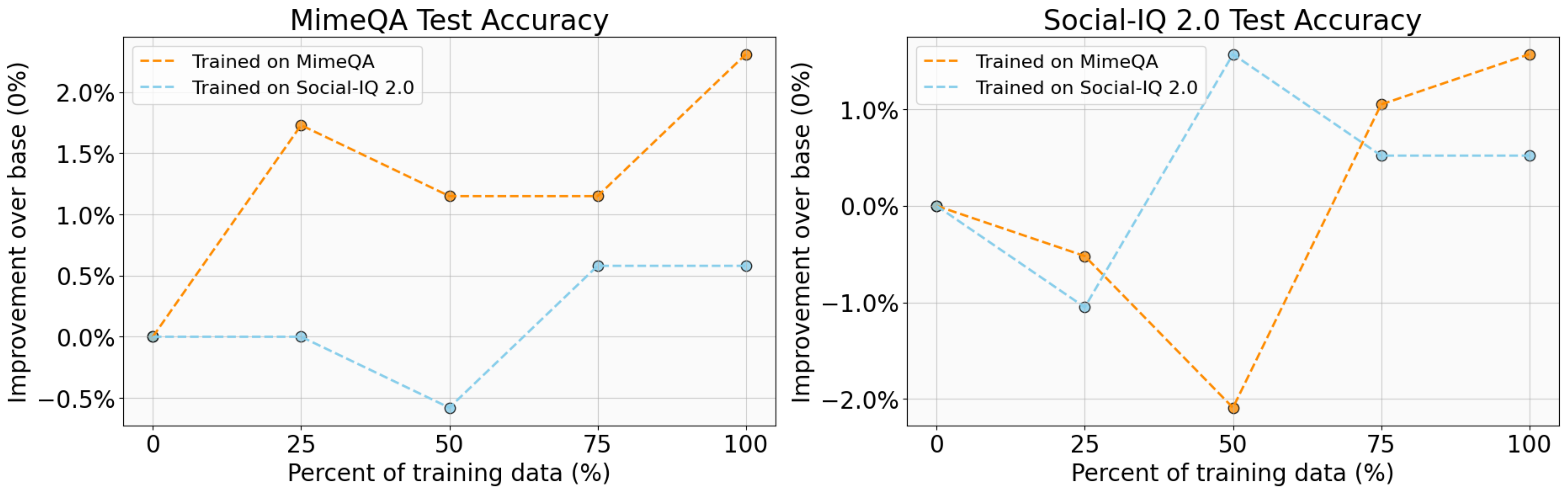}
    \caption{\textbf{Transfer analysis between \data\ and Social-IQ 2.0~\cite{siq2}.} Models fine-tuned on \data\ consistently generalize well to Social-IQ 2.0, while training on Social-IQ 2.0 yields little to no gains on \data. This highlights the distinct nonverbal social reasoning required in {\data} that is transferrable to other tasks.}
    \vspace{-6mm}
    \label{fig:mime_sft}
\end{figure}

The results, summarized in Table~\ref{tab:transfer_cv}, demonstrate the strong transferability of the skills learned from \data. On the Social-IQ 2.0 test, the model finetuned on \data\ achieves an average accuracy improvement of 1.2\%, which is comparable to the 1.0\% improvement from fine-tuning on Social-IQ 2.0 itself. Similarly, on the IntentQA test, the \data-finetuned model shows a 2.6\% improvement, only slightly lower than the 3.7\% gained from finetuning on IntentQA directly. These findings show that the nonverbal social understanding learned from \data\ effectively generalizes to broader social tasks. Conversely, models finetuned on Social-IQ 2.0 and IntentQA yielded much smaller accuracy improvements on \data\ at only 0.4\% and 1.1\%, respectively. Finetuning on \data\ offers significantly greater (p < 0.05) improvement of 3.5\% on its own test set, highlighting the distinct nonverbal reasoning required by our benchmark.

To further investigate these performance differences, we progressively finetuned \qwen\ on increasing subsets (25\%, 50\%, 75\%, 100\%) of the \data\ and Social-IQ 2.0 train sets. As shown in Figure \ref{fig:mime_sft}, increasing the training data from \data\ \textit{consistently} improves accuracy on its own test set, while training on Social-IQ 2.0 yields negligible gains. We hypothesize this gap occurs because the verbal nature of common real-world videos in Social-IQ 2.0 means they often lack the expressive body language central to mime performances. In fact, finetuning on Social-IQ  2.0 did lead to improvements on Theory of Mind (21.1\% to 27.6\%) and Intention (8.8\% to 11.7\%) questions, particularly on questions with sufficient grounding context. However, Social-IQ finetuning degraded performance in categories requiring precise interpretation of mimed actions: Temporal accuracy dropped from 9.1\% to 7.0\%, and Working Memory fell from 17.0\% to 14.2\%. This drop was often caused by increased hallucination. For example, when asked “Why did the person on the left lose what he was holding?”, the \data-tuned model correctly identified a quarrel, whereas the Social-IQ-tuned model misinterpreted the action as a dance routine. Thus, the social cues in Social-IQ 2.0 can be insufficient and even detrimental for tasks demanding robust nonverbal understanding. 

Overall, our results demonstrate that finetuning on \data\ yields consistent and transferable gains across diverse social reasoning benchmarks, highlighting its unique contribution of social information not captured by existing datasets.

\subsection{Error Analysis}

\begin{wrapfigure}{r}{0.55\textwidth}
\centering
\vspace{-10mm}
\includegraphics[width=0.65\linewidth]{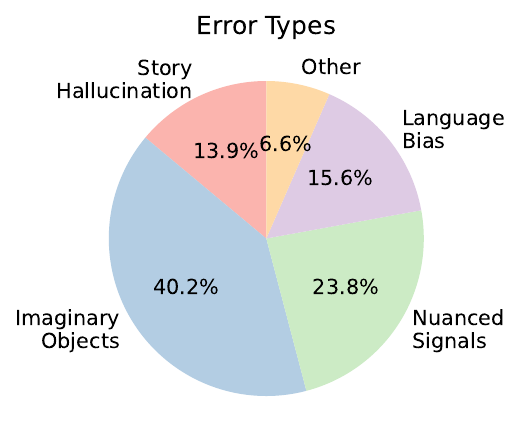}
\caption{\textbf{Error types distribution.} We annotate the error types for 20 videos and plot the distribution.}
\vspace{-8mm}
\label{fig:error_types}
\end{wrapfigure}

We highlight the main sources of errors by the VideoLLMs on \data, focusing on \gemini. We plot the distribution of sources of errors in Figure \ref{fig:error_types}.

\begin{figure}[t!]
    \centering
    \vspace{-4mm}
    \includegraphics[width=\linewidth]{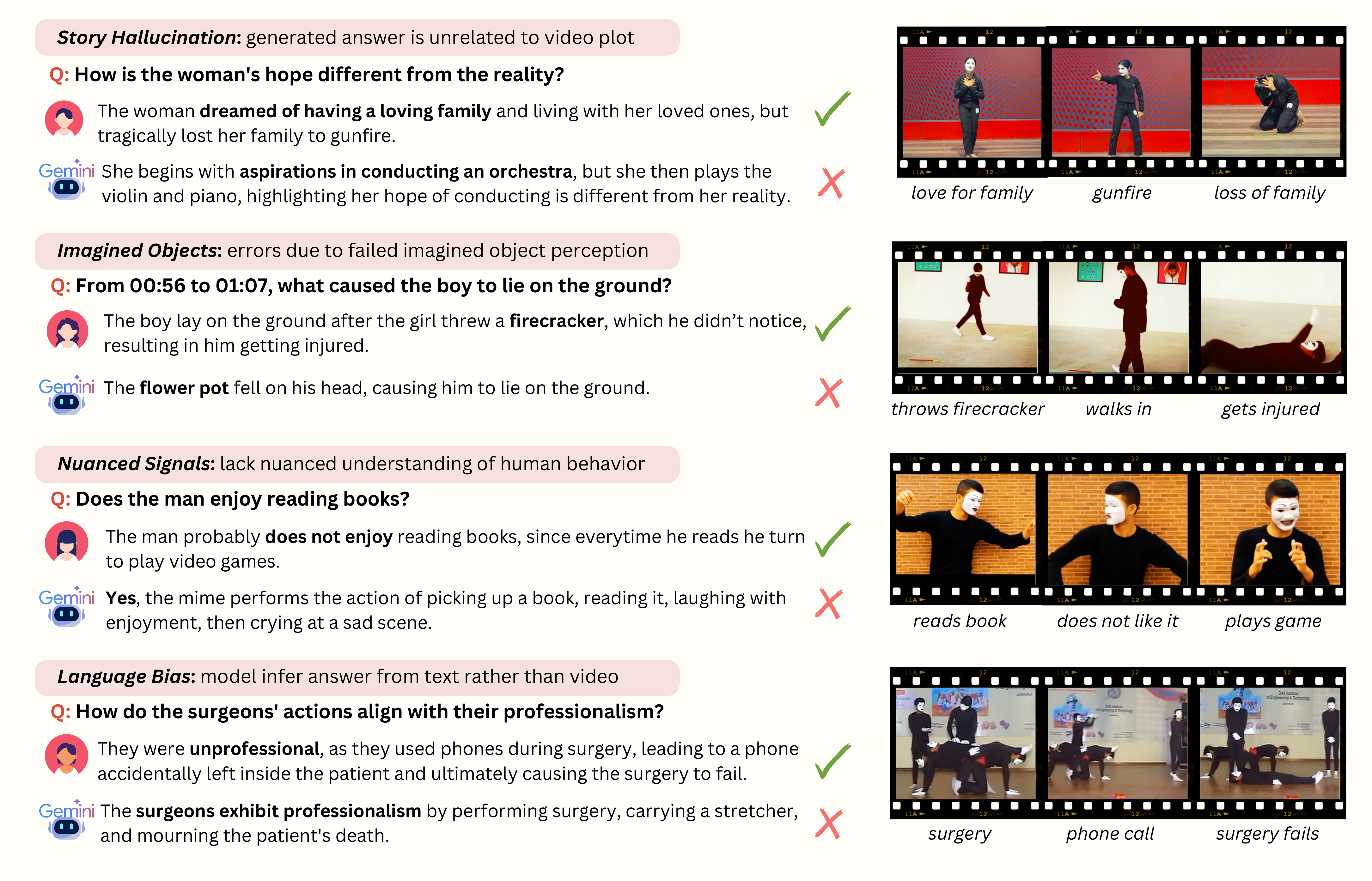}
    \caption{\textbf{{\data} model error examples.} We identify four common error categories. \textbf{Top-Bottom}. Story hallucination errors are when the model's response is unrelated to the video plot. Imagined objects denote errors where the model misidentifies the imagined objects. Nuanced signals denote instances of the model lacking a nuanced understanding of human behavior. Language bias denotes errors when the model is misled by the framing of the question and ignores the video.}
    \vspace{-6mm}
    \label{fig:error_examples}
\end{figure}

\paragraph{Story hallucination from missing language grounding.} One common pitfall pf today's best models on \data\ is hallucinating an answer that is plainly disconnected from the performance narrative. Due to the abstract and nonverbal nature of mime performances, VideoLLMs may interpret narratives in ways that deviate from commonsense. Figure \ref{fig:error_examples} contains an example where the mime is acting as a woman who, initially living peacefully with family, tragically lost her family during a war. However, {\gemini} misunderstands the narrative and hallucinates that the mime is conducting an orchestra, which is completely unrelated from the video.

\begin{wraptable}{r}{0.5\textwidth}
\centering
\vspace{-4mm}
\caption{\textbf{Model performance on videos with and without text.} Text in the video frames is detected automatically with manual verification. All models except for {\llava} have significantly improved performance on videos containing text.}
\scalebox{0.9}{
\begin{tabular}{ccc}
\toprule
\textbf{Model}& \textbf{With Text} & \textbf{Without Text}   \\
\hline
\qwen~\cite{qwen2.5-VL} &  \textbf{24.6} & 15.5 \\
\llava~\cite{zhangVideoInstructionTuning2024} & 19.2 & \textbf{19.5} \\
InternVL2.5~\cite{chenExpandingPerformanceBoundaries2025a} & \textbf{22.9} & 20.3 \\
VideoLLaMA3~\cite{zhangVideoLLaMA3Frontier2025} &  \textbf{27.1} & 17.3 \\
\hline
\gpt~\cite{achiam2023gpt} & \textbf{37.9} & 24.5 \\
\gemini~\cite{team2024gemini} & \textbf{35.2} & 26.0\\
\gemininew~\cite{comanici2025gemini} & \textbf{44.8} & 31.8 \\
\bottomrule
\end{tabular}
}
\vspace{-4mm}
\label{tab:models_text}
\end{wraptable}

We hypothesize that the model hallucinations stem from the lack of language grounding in mime performances, which provide no verbal context as in existing social video datasets with spoken communication. To test this hypothesis, we examined how model performance varies between videos containing meaningful text—such as hand-held signs or banners indicating the topic of the performance—and those without text. We sample frames from videos at one frame per second and use EasyOCR \citep{EasyOCR} for text detection. A human then verifies the detected text, filtering meaningless texts like watermarks. We report model accuracy on videos with and without text in Table \ref{tab:models_text}. We observe that most models, with the exception of \llava, achieved higher accuracy on videos containing text, which highlights their dependence on language modality and explains their poor performance on \data.

Additionally, we investigate whether providing video titles as supplementary language context improves model accuracy. We select \qwen~for experiment, and we report the results in Table \ref{tab:models_title}, where we observe that incorporating titles in the input prompt enhances accuracy across most categories. These results highlight a fundamental limitation: models heavily rely on language cues for social commonsense reasoning. To advance nonverbal social intelligence, we must \textbf{rethink visual cognition} in multimodal foundation models, ensuring better alignment of social signals across diverse modalities, especially when verbal information is not present.

\begin{table*}[h]
\centering
\caption{\textbf{Model performance with and without video title provided.}  T: text prompt includes title. \textbf{Avg}=Average performance across all questions. \textbf{GI}=Grounding the Imagined, \textbf{I}=Intention, \textbf{AR}=Affect Recognition, \textbf{T}=Temporal, \textbf{ToM}=Theory of Mind, \textbf{SJ}=Social Judgment, \textbf{WM}=Working Memory. {\qwen}'s performance improves across almost all categories when given the title.}
    \begin{tabular}{l c c ccc ccc}
    \toprule
    \multirow{2}{*}{\textbf{Model}} & \multirow{2}{*}{\textbf{Avg}} & \textbf{Grounding} & \multicolumn{3}{c}{\textbf{Scene-Level}} & \multicolumn{3}{c}{\textbf{Global-Level}} \\
    \cline{3-3} \cline{4-6} \cline{7-9}
    & & \textbf{GI} & \textbf{I} & \textbf{AR} & \textbf{T} & \textbf{ToM} & \textbf{SJ} & \textbf{WM} \\
    \hline
    \qwen~(with title) & \textbf{24.3} & \textbf{10.2} & \textbf{21.5} & 21.8 & \textbf{20.4} & \textbf{45.3} & \textbf{36.8} & \textbf{31.2} \\
    \qwen~(without title) & 20.1 & 6.6 & 15.8 & \textbf{23.6} & 14.3 & 38.7 & 33.3 & 19.4 \\
    \bottomrule
    \end{tabular}
\label{tab:models_title}
\end{table*}

\paragraph{Failure to interpret imagined objects.}
Understanding mime performances requires the audience to imagine invisible objects or activities from fine-grained gestures and body language \citep{sibierska2022s}. Our analysis suggests that models struggle to perceive imagined objects, leading to downstream reasoning errors in \data. For example, in Figure \ref{fig:error_examples}, a girl throws a firecracker on the ground, causing a boy to fall and appear injured. However, the model incorrectly identifies the firecracker as a flower pot. We also observe that the accuracy of Grounding is often positively correlated with correctness in other question categories (see Appendix \ref{app:grounding:corr}).

\begin{figure}[t!]
    \centering
    \vspace{-2mm}
    \includegraphics[width=0.9\linewidth]{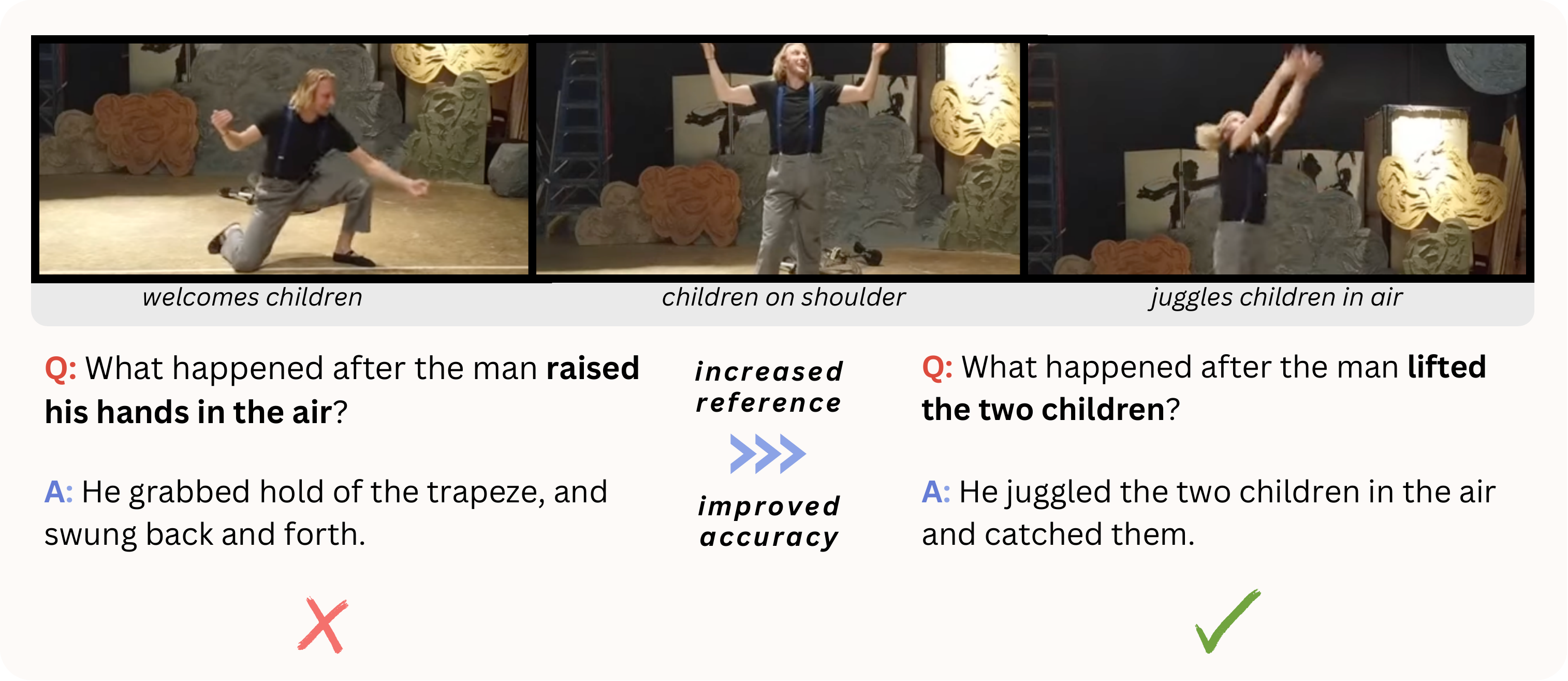}
    \caption{\textbf{Failure to interpret imagined objects impacts model's \data\ accuracy.} In this example, adding explicit reference to imagined objects in question allows {\gemini} to correctly answer the question.}
    \vspace{-2mm}
    \label{fig:incorrect_ref}
\end{figure}

To assess the impact of misperceived imagined objects on reasoning accuracy, we qualitatively analyze sample questions and examine how model responses change as object references become more explicit. We provide a representative example in Figure \ref{fig:incorrect_ref}. In this example, when initially asked what happens after the man in the video raises his hands, {\gemini} provides an incorrect response, misinterpreting the mime’s action as holding a trapeze. However, when the question is augmented with a clear description of the imagined objects, which are two children the man lifts onto his shoulders, {\gemini} now correctly responds that the mime is juggling them in the air. Building upon prior studies critically examining foundation models’ abstract visual cognition \citep{hsu2024makes, yiu2024kiva, schulze2025visual,su2025thinkingimagesmultimodalreasoning}, our findings highlight the need for \textbf{better human-AI perception alignment} \citep{muttenthaler2024aligning} to advance multimodal social intelligence.

\paragraph{Lack nuanced understanding of social signals.}
While models perform relatively well on Social Judgment and Theory of Mind compared to other categories, a closer examination reveals frequent errors stemming from a lack of nuanced understanding of human social signals. Figure \ref{fig:error_examples} illustrates such a case: a man begins reading a book but eventually loses interest and switches to playing a game. When asked whether the man enjoys reading, \gemini~ incorrectly responds affirmatively, relying on a naive interpretation of his initial reading behavior rather than recognizing his loss of interest. These global-level questions require models to integrate various local signals into a comprehensive narrative, highlighting the limitations of VideoLLMs in the complexity of social reasoning. Therefore, there is a need for research on \textbf{fine-grained social reasoning} which has been relatively understudied \citep{mathur-etal-2024-advancing, mathur2025social}.

\paragraph{Language bias over video content.}
Finally, we observe that models often infer answers based on the question prompt rather than the video content, making them prone to biases in language inputs. For example, in Figure \ref{fig:error_examples}, the mimes depict a scene where surgeons use their phones during surgery, accidentally leaving one inside the patient, resulting in their death. However, \gemini\ blindly and incorrectly identifies the surgeons as acting professionally, relying on prior assumptions and biases from its language pretraining rather than interpreting the visual narrative. This analysis is further supported by models' text-only accuracy results in Table \ref{tab:models} which show that, particularly for open-source models, performance improves only marginally when video context is provided alongside the question text. This suggests that VideoLLMs' reliance on the question prompt rather than genuine video understanding.

The above findings underscore the need for multimodal models that effectively integrate all input modalities on language rather than over-rely on language, as also observed in other works \citep{fu2025hidden}. Additionally, while social bias in language models has been widely studied \citep{liang2021towards, gallegos2024bias}, our results emphasize the need to \textbf{understand and mitigate how these biases transfer} to multimodal social tasks, given the models' dependence on language.
\section{Conclusion}
Our \data\ benchmark highlights the crucial need for video LLMs to move beyond linguistic bias by integrating deeper non-verbal understanding for socially intelligent AI. By proposing mime understanding as a novel evaluation setting, we introduce a challenging yet valuable benchmark that requires models to interpret human gestures, emotional dynamics, and social interactions without explicit spoken dialogue. Our fine-tuning experiments reveal that while targeted training can improve higher-level social reasoning capabilities, fundamental challenges in grounding imagined objects persist. The asymmetric transfer patterns between \data\ and an existing social benchmark demonstrate that \data\ captures distinct aspects of social cognition, with mime-trained models generalizing well while models trained on conventional social datasets show minimal improvement on mime understanding. Our comprehensive analysis presents new research directions toward advancing the next generation of verbal and nonverbal socially intelligent foundation models.

\newpage

\section*{Acknowledgement}
MT is supported by the National Science Foundation (NSF) under Grant No. 2141064. Any opinion, findings, and conclusions or recommendations expressed in this material are those of the authors and do not reflect the views of the National Science Foundation. We thank the MIT Office of Research Computing and Data (ORCD) and the NVIDIA Academic Grant Program for GPU support.

{\footnotesize
\bibliographystyle{plainnat}
\bibliography{refs}
}



\newpage

\newpage
\appendix
\onecolumn

\section{Limitations, Ethics Statement and Broader Impact}\label{app:limitations}
In this paper, we focused on benchmarking AI models beyond linguistic cues by exploring the domain of mimes -- a uniquely rich form of nonverbal, embodied human communication. However, there are many other expressive, human-centered interaction modalities that remain underexplored and present promising avenues for future research, including interactive art installations, dance, cultural performances, and musical expressions. These domains offer distinct affordances for evaluating and enhancing AI’s capacity for social understanding. Additionally, our current dataset is limited in size, and developing effective training methods that can achieve robust social reasoning from small data regimes remains an open challenge. Finally, the cultural and demographic contexts reflected in our benchmark primarily represent Western societies, and semantic subtleties vary across cultures \cite{huang2025cultureclipempoweringclipcultural}. A crucial direction for future work is to broaden the cultural scope of \data\ to encompass a more diverse range of social norms, behaviors, and modes of expression across global communities.

All human annotations and verifications, both for dataset construction and analysis, were conducted by the authors. Details of the annotation instructions can be found in Appendix \ref{app:annotation_details}. All video data used in this article are publicly available under Creative Commons license, and none of the annotations contain personally identifiable information. {\data} is released under a CC-BY-NC-SA 4.0 license and is intended solely for research purposes.

Developing rich nonverbal social reasoning is crucial for AI systems to interact effectively with humans and enhance well-being. While our work aims to advance this capability, we acknowledge the potential risks associated with such advancements, including applications in mass surveillance that could infringe on individual privacy. We support efforts to mitigate potential misuse and to ensure that socially intelligent AI systems are developed and applied responsibly.

\section{Additional Results}
\subsection{Diversity Analysis of \data}\label{app:diversity}
We acknowledge that the interpretation of mime performances may be influenced by cultural context and perceived gender of the performer. To understand the cultural and demographic biases in \data, we randomly sampled 25\% of the videos in the dataset and manually annotated two attributes using both the video content and its title/description: 1) whether the cultural context appeared Western or non-Western, and 2) the perceived binary genders (male/female) of the individuals depicted. Our analysis indicates that 56\% of the sampled videos reflect non-Western cultural contexts, with many sources originating from Asian regions. In terms of gender, 88\% of videos feature at least one male participant, and 68\% include at least one female participant. A majority of videos contain both male and female participants. Thus, although \data\ was not explicitly curated to enforce cultural or gender diversity, these findings suggest a reasonably balanced representation across both dimensions, with no dominant cultural or gender skew.

\begin{wrapfigure}{r}{0.5\textwidth}
    \centering
    \includegraphics[width=\linewidth]{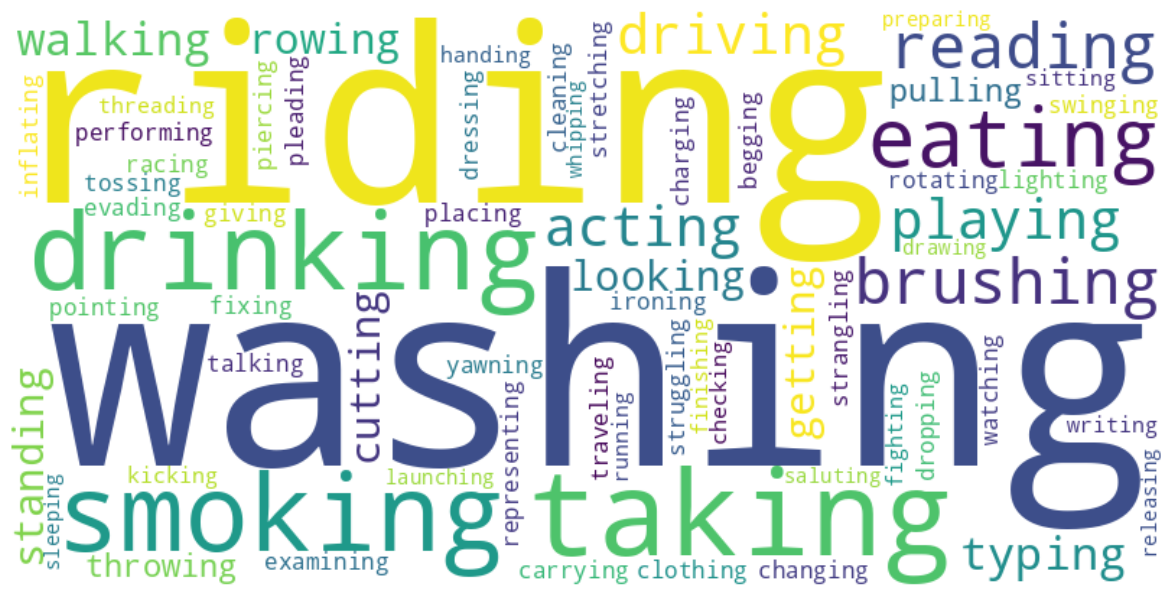}
    \caption{\textbf{Word cloud of actions in \data.} There are 68 distinct action keywords in \data's answers, indicating its human behavior diversity.}
    \label{fig:action_keywords}
\end{wrapfigure}

We further elaborate on the diversity of non-verbal social cues in \data. We examined the reference answers for the “Grounding the Imagined” category and extracted all words ending in “-ing,” yielding a list of action-related terms. After removing mime- or performance-specific words (e.g., “pretending,” “portraying”), we obtained 68 unique action keywords, comparable to the range found in many action recognition datasets. We provide a word cloud of the action keywords in Figure \ref{fig:dataset_pipeline}. To further estimate the diversity of actions in \data, we ran PoseC3D \citep{duan2022revisiting}, an action recognition model trained on the NTU RGB+D dataset \citep{shahroudy2016ntu}, on 10-second clips from a 30\% subset of \data\ videos. Out of 60 possible action categories, the model identified 46 distinct actions across the subset. While this is only a proxy measure, we believe it reflects the broad range of perceived nonverbal behavior captured in \data.

\subsection{Does Chain-of-Thought Improve Accuracy on \data?}\label{app:cot}
We additionally experiment whether model accuracy on {\data} could be improved with chain-of-thought (CoT) prompting \citep{wei2022chain}. To integrate CoT prompting, we add the following line to the end of evaluation prompt. 
\begin{quotation}
\texttt{"Think step by step to answer the question. Format the final answer in a separate sentence like 'The answer is X'."}
\end{quotation}

We report the results in Table \ref{tab:cot}. Overall, although CoT prompting slightly improves \qwen's accuracy on \data, for most models it slightly decreases performance. Taking a closer look, we find that, for example, \gpt\ with CoT prompting tends to hallucinate more for the Grounding the Imagined, Temporal Reasoning, and Affect Recognition categories, thereby leading to performance drop. This finding is in line with prior work \citep{spraguecot} that shows CoT is not always helpful for tasks which lack symbolic reasoning. 

\begin{table}[h]
    \centering
    \caption{\textbf{Model performance on \data\ with and without chain-of-thought (CoT).} We find that CoT yields no significant accuracy improvement on \data.}
    \vspace{2mm}
    \begin{tabular}{lcc}
        \toprule
        \textbf{Model} & \textbf{Without CoT} & \textbf{With CoT} \\
        \midrule
        \qwen & 20.1\% & \textbf{22.2\%} \\
        \llava & \textbf{19.4\%} & 17.7\% \\
        \internvl & \textbf{21.6\%} & 18.7\% \\
        \videollama & \textbf{22.2\%} & 17.7\% \\
        \gpt & \textbf{31.3\%} & 29.5\% \\
        \bottomrule
    \end{tabular}
     \label{tab:cot}
\end{table}

\subsection{Correlation between Grounding and Other Question Categories}\label{app:grounding:corr}

To analyze the effect of the model's inability to understand localized events, we compute the correlation between performance on Grounding the Imagined questions to the other scene-level and global-level questions. Intuitively, we would expect that a model's ability to perform grounding would correlate strongly with, for example, temporal understanding, as one needs to understand individual actions and objects before reasoning about a sequence of events. 

\begin{figure}[h!]
    \vspace{-4mm}
    \centering
    \includegraphics[width=0.44\linewidth]{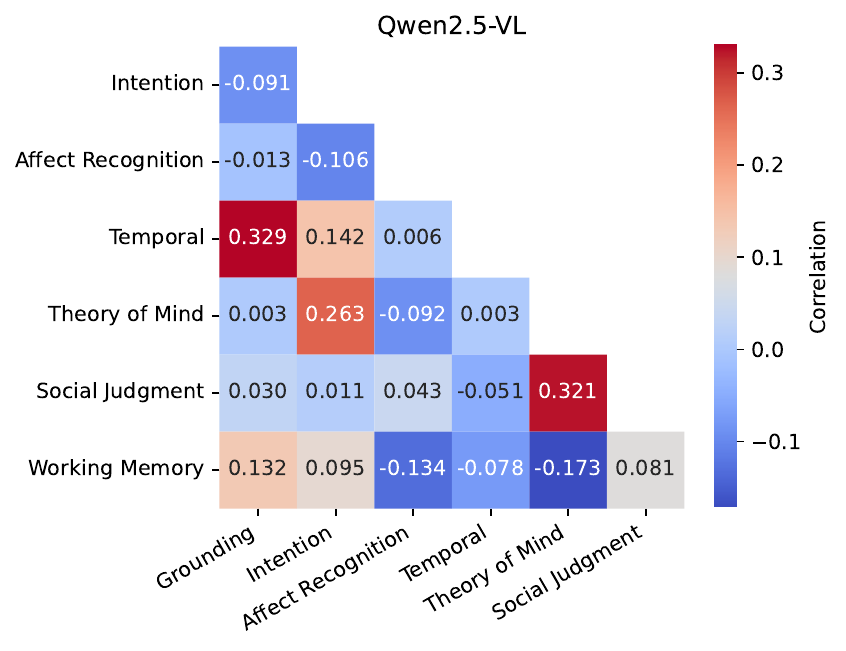}
    \includegraphics[width=0.44\linewidth]{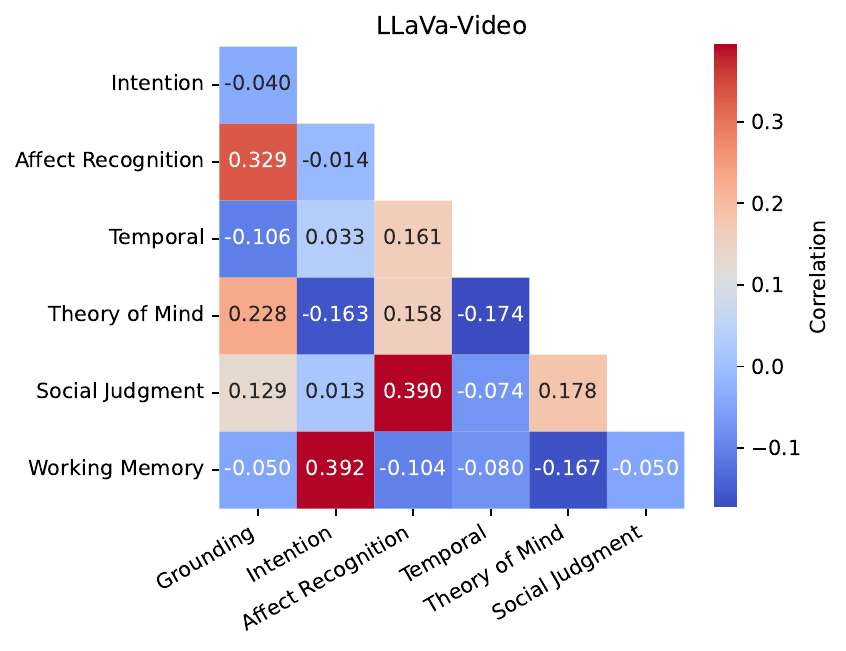} \\
    \includegraphics[width=0.44\linewidth]{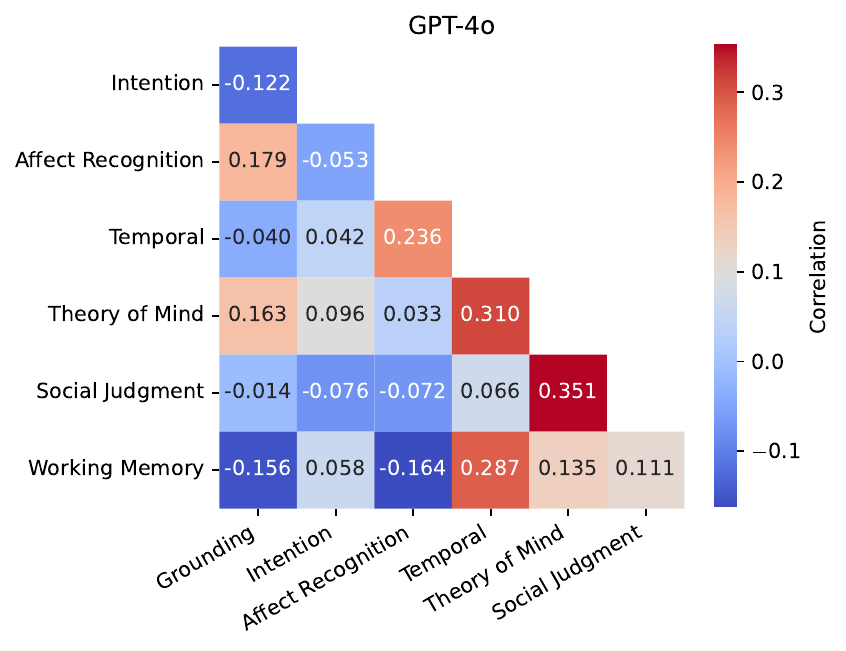}
    \includegraphics[width=0.44\linewidth]{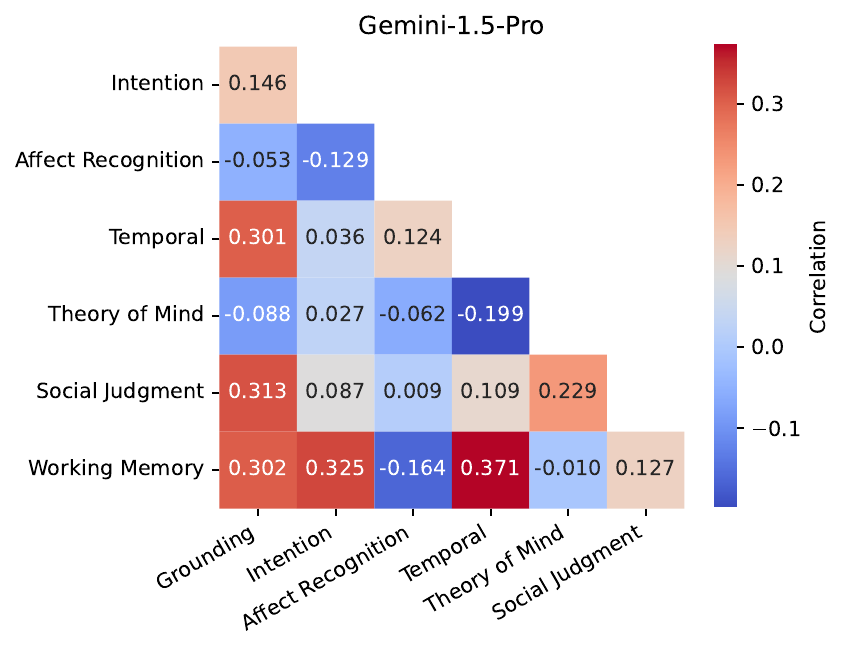}
    \caption{\textbf{Question type performance correlation matrices on video and text input.} For each video, we compute the accuracy over all question types and plot the correlation between accuracies on different question types. From left to right, we show the correlation matrices for {\qwen}, {\llava}, {\gemini}, and {\gpt}.   }
    \vspace{-4mm}
    \label{fig:vl_corr}
\end{figure}

\begin{figure}[h!]
    \vspace{-4mm}
    \centering
    \includegraphics[width=0.44\linewidth]{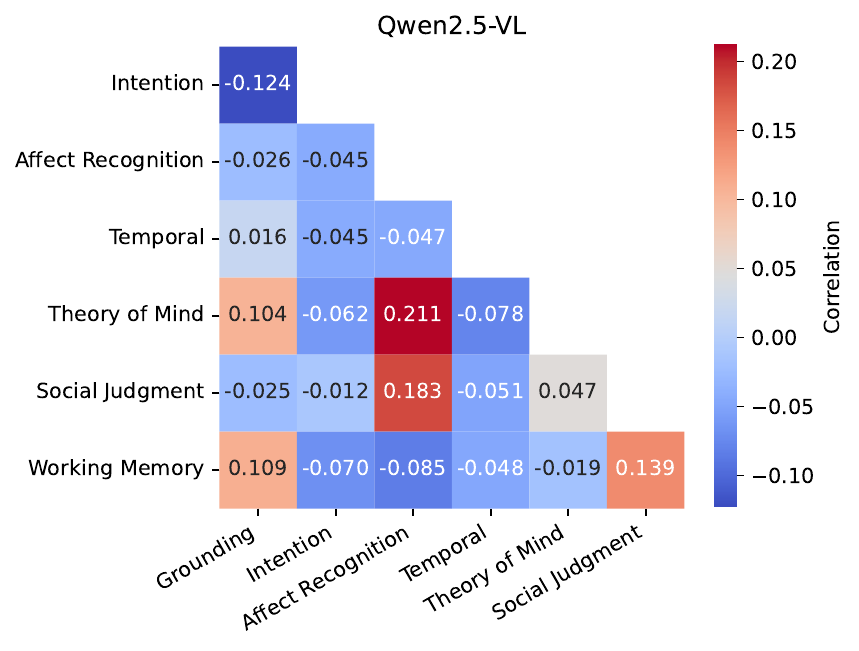}
    \includegraphics[width=0.44\linewidth]{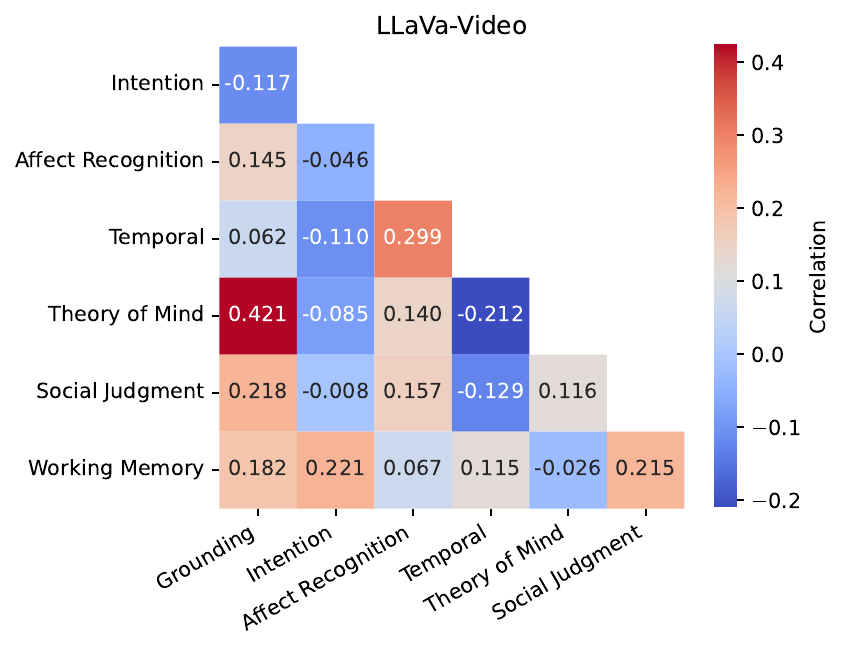}
    \includegraphics[width=0.44\linewidth]{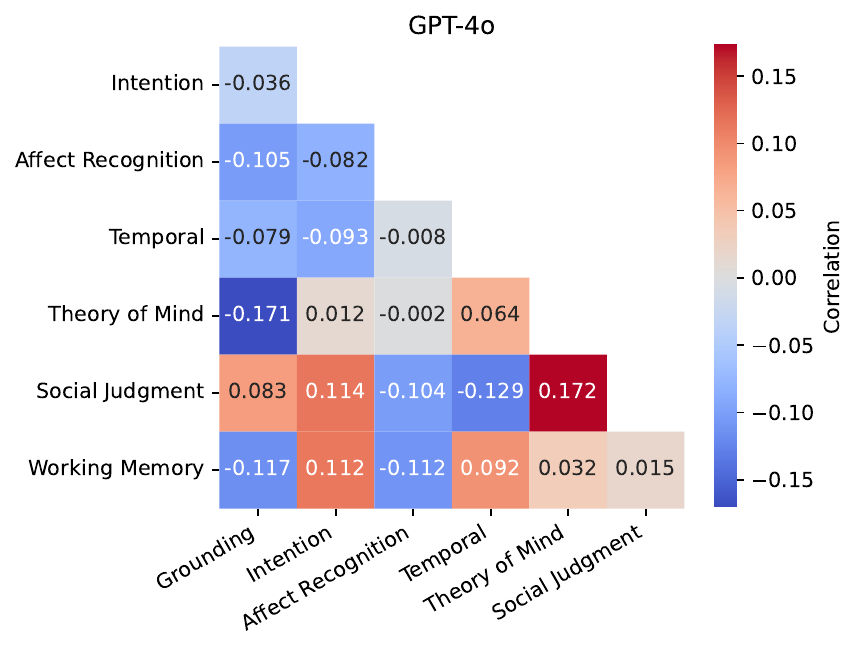}
    \includegraphics[width=0.44\linewidth]{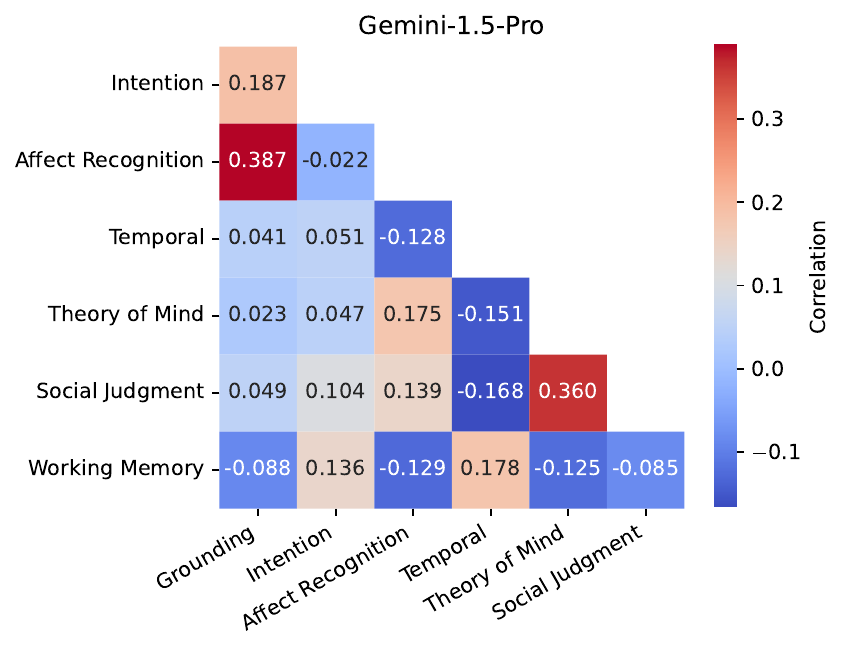}
    \caption{\textbf{Question type performance correlation matrices on only text input.} To ablate the effect of video, we perform inference for all models with only the text prompt. For each video, we compute the accuracy over all question types and plot the correlation between accuracies on different question types. From left to right, we show the correlation matrices for {\qwen}, {\llava}, {\gemini}, and {\gpt}.   }
    \vspace{-4mm}
    \label{fig:l_corr}
\end{figure}

In Table \ref{tab:models_corr}, we report the computed accuracy correlation for selected models. We find that \qwen's Grounding performance positively correlates with Temporal understanding. For \llava, Grounding accuracy correlates with Affect Recognition and Theory of Mind. For \gemini, we see that Grounding performance contributes both to understanding localized temporal sequences as well as to a more holistic understanding of the video, as shown by higher correlation scores with Temporal understanding, Social Judgment, and Working Memory. For \gpt, Grounding performance correlates with Affect Recognition and Theory of Mind. See Figure \ref{fig:vl_corr} for correlation between all pairs of question categories when the input contains both video and language, and Figure \ref{fig:l_corr} for all correlations when the input is language only. This suggests that improved understanding of the fine-grained visual cues would lead to a better grasp of the video plot, with the specific reasoning pathways that benefit from this grounding being model-dependent. Our results demonstrate that an important line of future work is to improve VideoLLMs' ability to reason without explicit objects or human-object interactions, which can bottleneck performance on holistic video understanding. 

\begin{table}[h]
\centering
\caption{\textbf{Performance correlation between grounding the imagined questions and other categories for selected models.} \textbf{I}=Intention, \textbf{AR}=Affect Recognition, \textbf{T}=Temporal, \textbf{ToM}=Theory of Mind, \textbf{SJ}=Social Judgment, \textbf{WM}=Working Memory. }
\vspace{2mm}
\begin{tabular}{ccccccc}
\toprule
\textbf{Model}& I & AR & T & ToM & SJ & WM   \\
\hline
\qwen &  -0.091 & -0.013 & 0.330 & 0.003 & 0.030  & 0.132 \\
\llava &  -0.040 & 0.329 & -0.106 & 0.228 & 0.129 & -0.050 \\
\hline
\gpt & -0.122& 0.179 & -0.040 & 0.163 & -0.014 & -0.156\\
\gemini & 0.146 & -0.053 & 0.301 & -0.088 & 0.313 & 0.302 \\
\bottomrule
\end{tabular}
\vspace{2mm}
\label{tab:models_corr}
\end{table}

\subsection{How Does Reasoning Vision-Language Models Perform on \data?}
We evaluated the performance of frontier vision-language models not specifically fine-tuned for video understanding, focusing on OpenAI's GPT-o3 \citep{OpenAI2025o3} on the \data\ benchmark. Following a similar setup to \gpt, videos were sampled at 1 frame per second to a maximum of 64 frames. The results, presented in Table \ref{tab:gpt-o3}, show that GPT-o3 (34.4) demonstrates exceptional strength in global-level questions requiring sophisticated reasoning. It significantly outperforms all other models on Theory of Mind (62.7) and Working Memory (41.6). It also performs well on scene-level Intention (32.3) and Temporal (34.7) reasoning.

However, these advanced reasoning capabilities appear to come at the cost of visual perception. GPT-o3 struggles with perception-heavy tasks, lagging behind {\gemininew} in Affect Recognition (32.3 vs. 43.7). Notably, it scores just 13.1 on Grounding the Imagined, falling significantly behind both \gpt\ (19.0) and \gemini\ (28.4). This highlights a potential trade-off where language reasoning gains may compromise foundational visual grounding capabilities.

\begin{table}[h]
\centering
\vspace{-4mm}
\caption{\textbf{GPT-o3 accuracy on \data\ compared to selected models.} GPT-o3 shows the most significant gains on global level questions.}
\vspace{2mm}
\scalebox{0.85}{
\begin{tabular}{l c | c | c c c | c c c}
\toprule
\textbf{Model} & \textbf{Avg} 
& \multicolumn{1}{c|}{\textbf{Grounding}} 
& \multicolumn{3}{c|}{\textbf{Scene-Level}} 
& \multicolumn{3}{c}{\textbf{Global-Level}} \\
\cmidrule(lr){3-3} \cmidrule(lr){4-6} \cmidrule(lr){7-9}
& & \textbf{GI} & \textbf{I} 
& \textbf{AR} & \textbf{T} & 
\textbf{ToM} & \textbf{SJ} & \textbf{WM} \\
\midrule
\gpt &  31.3 & 19.0 & 28.5  & 29.9 & 30.6 & 45.3 & 43.7 & 35.1 \\
\gemininew & \textbf{38.3} & \textbf{28.4} & 31.6 & \textbf{43.7} & 28.6 & 54.7 & \textbf{51.7} & 39.0 \\
GPT-o3 & 34.4 & 13.1 & \textbf{32.3} & 32.2 & \textbf{34.7} & \textbf{62.7} & 44.8 & \textbf{41.6} \\
\hline
Human & 86.0 & 89.8 & 87.3 & 83.9 & 88.8 & 93.3 & 80.5 &  76.6 \\
\bottomrule
\end{tabular}
}
\vspace{-2mm}
\label{tab:gpt-o3}
\end{table}


\section{Additional Experimental Details}\label{app:exp_details}
\subsection{Video Collection Details}
We searched YouTube for videos using eight keywords: \texttt{mime performance}, \texttt{mime act}, \texttt{mime artist}, \texttt{mime comedy}, \texttt{mime theatre}, \texttt{mime play}, \texttt{mime skit}, and \texttt{mime video}. For each search, we collected a maximum of 50 results, keeping only those videos that are between 1 and 10 minutes in duration and released under a Creative Commons license.

\subsection{Model Settings and Parameters}\label{app:model_settings}
We set the maximum output tokens for each model to be 128 tokens. We detail the settings of the models below. 
\begin{itemize}[noitemsep,topsep=0pt,nosep,leftmargin=*,parsep=0pt,partopsep=0pt]
    \item \gemini~\citep{team2024gemini} and \gemininew~\citep{comanici2025gemini}: natively supports video as input, including audio. 
    \item \gpt~\citep{achiam2023gpt}: We sample 1 frame per second up to a maximum of 64 frames, in which case the frames are uniformly sampled. We resize the image to 512x512 to fit in the context window.
    \item Qwen2.5-VL-72B \citep{qwen2.5-VL}: natively supports video as input, sampled at 2 frames per second for a maximum of 768 frames.
    \item LLaVA-Video-72B \citep{zhangVideoInstructionTuning2024}: We sample 1 frame per second up to a maximum of 384 frames, in which case the frames are uniformly sampled.
    \item \internvl~\citep{chenExpandingPerformanceBoundaries2025a}: natively supports videos as input.
    \item \videollama~\citep{zhangVideoLLaMA3Frontier2025}: natively supports video as input, with a maximum of 180 frames.
\end{itemize}

\subsection{Supervised Finetuning}
All fine-tuning experiments were conducted using the LLaMA Factory framework \citep{zheng2024llamafactory}. For the Qwen2.5-VL-72B model, we applied QLoRA fine-tuning with 8-bit quantization, a LoRA rank of 8, a learning rate of $1 \times 10^{-4}$, and trained for 3 epochs on \data. For the Qwen2.5-VL-7B model, we used standard LoRA fine-tuning with a rank of 8, a learning rate of $3 \times 10^{-6}$, and trained for 2 epochs. No additional hyperparameter tuning was performed.

\subsection{Compute Resources}\label{app:compute}
All evaluations and experiments in this paper were conducted on a remote cluster equipped with two NVIDIA H200 GPUs (each with 141 GB HBM3 memory). Runtime for each model evaluation varied between 1–10 hours depending on the model size.

\subsection{Prompt Details}\label{app:prompt}

Below is the prompt template for {\gemini}, {\gemininew}, and {\qwen}, which natively take in video input.
\begin{lstlisting}[style=markdownstyle]
You are an expert in mime performance understanding and question answering. 
Typically, the mime would use exaggerated gestures or pretend objects to convey a message.
Answer the question in one sentence using the video, with brief explanations. 
Do not describe the frames just answer the question, and say nothing else.
If the mime is using imaginary objects, describe the objects as if they were real.
Question: <question>
\end{lstlisting}

As {\gpt} and {\llava} require frames to be sampled from the video, we additionally specify the length of the video and the timestamps of the sampled frames in the text prompt. Below is the prompt template for {\gpt} and {\llava}.

\begin{lstlisting}[style=markdownstyle]
You are an expert in mime performance understanding and question answering. 
Typically, the mime would use exaggerated gestures or pretend objects to convey a message.
The video lasts for {video_time}, and {num_frames} frames are uniformly sampled from it.
These frames are located at {frame_time}. Answer the question in one sentence using the video, with brief explanations. 
Do not describe the frames just answer the question. 
If the mime is using imaginary objects, describe the objects as if they were real.
Question: <question>
\end{lstlisting}

Below is the prompt to {\gpt} for LLM-as-a-judge, which is adapted from \citet{nagrani2024neptune}.

\begin{lstlisting}[style=markdownstyle]
Answer Grading Instructions:
Carefully consider the following question and answers regarding understanding of a mime performance.
You will be shown a "gold-standard" answer from a human annotator, referred to as the "Reference Answer", and a "Candidate Answer".
Your task is to determine whether the candidate answer is a good answer in place of the "gold" reference using the following criteria:

1. The candidate directly answers the question without deviation or misunderstanding.
2. The candidate does not contain misleading information and does not hallucinate story plots not present in the reference answer.
3. Since the videos are mime performances, invisible actions, objects, or the mime actor portraying objects should be considered correct if and only if they are relevant to the question.
4. The candidate answer can be a good answer in place of the reference answer even if they are not semantically equivalent, as long as they are in the same ballpark, given that there can be multiple interpretations of a mime acting out invisible actions or objects.

Your response should be one word, "TRUE" or "FALSE", and a brief explanation of your decision. You should respond "TRUE" if the candidate is a good answer in place of the reference answer, and "FALSE" otherwise.
"""

GRADE_PROMPT = """
Question:
"{question}"
Candidate Answer:
"{candidate_answer}"
Reference Answer:
"{ref_answer}"
Equivalent? (True/False, and why?):
\end{lstlisting}

\section{Human Annotation Details}\label{app:annotation_details}
\subsection{Guidelines for Annotators}

\begin{tcolorbox}[colframe=black, colback=white, arc=3mm, boxrule=1pt, width=\linewidth, title=\textbf{Annotation Instructions}, breakable]
Each video ranges from 1 minute to 10 minutes. For each video, aim for approximately 6 scene-level questions, 4 global-level questions, and any relevant grounding questions. Mark explicit START and END timeframes for shot-level and local-level questions. Here is the question hiearchy:

\textbf{Grounding} This level mainly serves as a sanity check for whether the model understands the portrayed object in mime videos – what’s the person doing, holding, and describing the imagined objects depicted.
    \begin{itemize}
        \item \textit{E.g. What is the person in black shirt doing/holding/etc.?}
    \end{itemize}

\textbf{Scene-Level} — local social information and temporal connection
\begin{itemize}
    \item \textbf{Temporal} Interpreting sequences of events, causality, and the flow of actions within a specific timeframe.
    \begin{itemize}
        \item \textbf{Template:} What caused (some event) to happen? 
        \item \textit{E.g. What caused the person to fall over?}
        \item \textit{E.g. What happened before the person placed the spoon on the table?}
    \end{itemize}

    \item \textbf{Affect Recognition} Tracking and analyzing emotions within a local scene, including subtle transitions and group sentiment. 
    \begin{itemize}
        \item \textbf{Template:} What is the attitude of (some person) towards (some event)? 
        \item \textbf{Template:} How does the (person) feel when (some event) happened?
        \item \textit{E.g. How is the person in black shirt feeling after placing the stone?} 
        \item \textit{E.g. What is the attitude of the man towards the woman?}
        \item \textit{E.g. How did the group's emotional tone shift during the interaction?}
    \end{itemize}

    \item \textbf{Intention and Behavior} Interpreting goals, plans, and motivations within a scene. 
    \begin{itemize}
        \item \textbf{Template:} Why did the (person) do (some action)?
        \item \textit{E.g. Why did the person holding the ice cream cry?}
        \item \textit{E.g. Why is the person in black outfit not speaking?}
        \item \textit{E.g. Why did the woman pretend not to notice the man?}
        \item textit{E.g. Why did the individual wait their turn before speaking?}
    \end{itemize}
\end{itemize}

\textbf{Global-Level} — focus on the overall narrative and high-level concepts
    \begin{itemize}
        \item \textbf{Working Memory} Retrieving, integrating, and reasoning with information across the entire video, beyond localized linear sequences. Requires the ability to decide relevance of information and present a coherent narrative.
        \begin{itemize}
            \item \textbf{Template:} What happened after (some event)? 
            \item \textbf{Template:} How has the relationship between (person) and (person) changed?
            \item \textbf{Template:} What would happen if (an event) didn’t happen? 
            \item \textit{E.g. What object in the beginning of the video foreshadowed the outcome?}
            \item \textit{E.g. How has the actions of the person changed throughout the video?}
            \item \textit{E.g. What event in the start of the video triggered the conflict in the final scene?}
        \end{itemize}
        \item \textbf{Social Judgment} Evaluating behaviors, morality, and adherence to social norms, with consideration for cultural context and moral reasoning.
        \begin{itemize}
            \item \textbf{Template:} How are the (person) and (person) getting along?
            \item \textbf{Template:} How do the (person) actions demonstrate (social concept)?
            \item \textit{E.g. How does the person in the black shirt demonstrate rapport with the person in the blue dress?}
            \item \textit{E.g. What do the person’s actions tell about his personality?}
            \item \textit{E.g. How might the group perceive this individual’s behaviour?}
            \item \textit{E.g. How do the characters’ behaviors suggest they are cooperating?}
        \end{itemize}

        \item \textbf{Perspective Taking (Theory of Mind)} Inferring beliefs, desires, and emotions of others, including both cognitive and affective understanding.
        \begin{itemize}
            \item \textbf{Template:} Does (person A) understand what (person B) was feeling?
            \item \textbf{Template:} What is the (person) hoping to achieve?
            \item \textbf{Template:} Would (person) do (action) after (event)? 
            \item \textit{E.g. What goal does the main character pursue throughout the video?}
            \item \textit{E.g. How is the character’s hope different from the reality?}
            \item \textit{E.g. Why is the main character motivated to change his behavior?}
        \end{itemize}
    \end{itemize}
\end{tcolorbox}

\subsection{Guidelines for Verifiers}

\begin{tcolorbox}[colframe=black, colback=white, arc=3mm, boxrule=1pt, width=\linewidth, title=\textbf{Verification Instructions}, breakable]
This is a video question-answering dataset consisting of mime videos. The goal of this dataset is to evaluate whether current video language models can perform rich visual social reasoning without relying on natural language. 
\begin{itemize}
    \item Watch the entire video before reviewing the questions. The videos can be found [\textit{link}]
    \item Answer each question based on the video content. If a question refers to a specific timestamp, focus on that section; otherwise, consider the whole video.
    \item Compare your answer with the “Reference Answer” column: Mark T in the “Answer Aligned” column if they align. ark F if they are clearly misaligned.
    \item For ambiguous questions, suggest a clearer version in the “Alternative Question” column.
\end{itemize}
\end{tcolorbox}

\end{document}